\RequirePackage{fix-cm}
\documentclass[twocolumn]{svjour3}          % twocolumn
\smartqed  % flush right qed marks, e.g. at end of proof

\usepackage{cite}
\usepackage[pdftex]{graphicx}
\usepackage{ragged2e}
\usepackage[tight,footnotesize]{subfigure}
\usepackage{rotating}
\usepackage{graphicx}
\usepackage{amsmath,amssymb} % define this before the line numbering.
\usepackage{array}
\usepackage{multirow}
\usepackage{colortbl}
\usepackage{amsfonts}
\usepackage{pifont}
\usepackage{xspace}
\usepackage{etoolbox}
\usepackage{overpic}
\usepackage{color}
\usepackage{microtype}
\usepackage{booktabs}

\newcommand{\orcid}[1]{\href{https://orcid.org/#1}{\includegraphics[width=10pt]{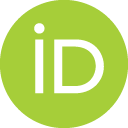}}}

\def\etal{{\em et al}}

\usepackage{hyperref}
\hypersetup{breaklinks=true,citecolor=blue, colorlinks}

\graphicspath{{./Imgs/}}
\DeclareGraphicsExtensions{.pdf,.jpg,.png}

\usepackage{silence}
\journalname{Research Article}

\begin{document}

\title{Recent Advances of Continual Learning in Computer Vision: An Overview}

\titlerunning{Recent Advances of Continual Learning in Computer Vision: An Overview}        % For running head

\author{Haoxuan Qu \orcid{0000-0001-5054-3394}        \and
  Hossein Rahmani \orcid{0000-0003-1920-0371} \and 
  Li Xu \orcid{0000-0003-1575-5724} \and
  Bryan Williams \orcid{0000-0001-5930-287X} \and
  Jun Liu \orcid{0000-0002-4365-4165}
}

\authorrunning{Qu \etal} % if too long for running head

\institute{
Haoxuan Qu, Hossein Rahmani, Bryan Williams, and Jun Liu are with Lancaster University, Lancaster, Lancashire, UK.  \\
Li Xu is with Singapore University of Technology and Design, Singapore, Singapore. \\
Corresponding author: Jun Liu.
}

\date{Received: date / Accepted: date}

\maketitle

\begin{abstract}
In contrast to batch learning where all training data is available at once, continual learning represents a family of methods that accumulate knowledge and learn continuously with data available in sequential order. Similar to the human learning process with the ability of learning, fusing, and accumulating new knowledge coming at different time steps, continual learning is considered to have high practical significance. Hence, continual learning has been studied in various artificial intelligence tasks. In this paper, we present a comprehensive review of the recent progress of continual learning in computer vision. In particular, the works are grouped by their representative techniques, including regularization, knowledge distillation, memory, generative replay, parameter isolation, and a combination of the above techniques. For each category of these techniques, both its characteristics and applications in computer vision are presented. At the end of this overview, several subareas, where continuous knowledge accumulation is potentially helpful while continual learning has not been well studied, are discussed.

% Please provide 4 to 6 keywords which can be used for indexing purposes.

\end{abstract}

\section{Introduction}

Human learning is a gradual process. Throughout the course of human life, humans continually receive and learn new knowledge. While new knowledge plays a role in its own accumulation, it also supplements and revises previous knowledge. In contrast, traditional machine learning and deep learning paradigms generally distinguish the processes of knowledge training and knowledge inference, where the model is required to complete its training on a pre-prepared dataset within a limited time which can then be used for inference. 

\begin{figure*}[t]
\centering
  \includegraphics[width=\linewidth]{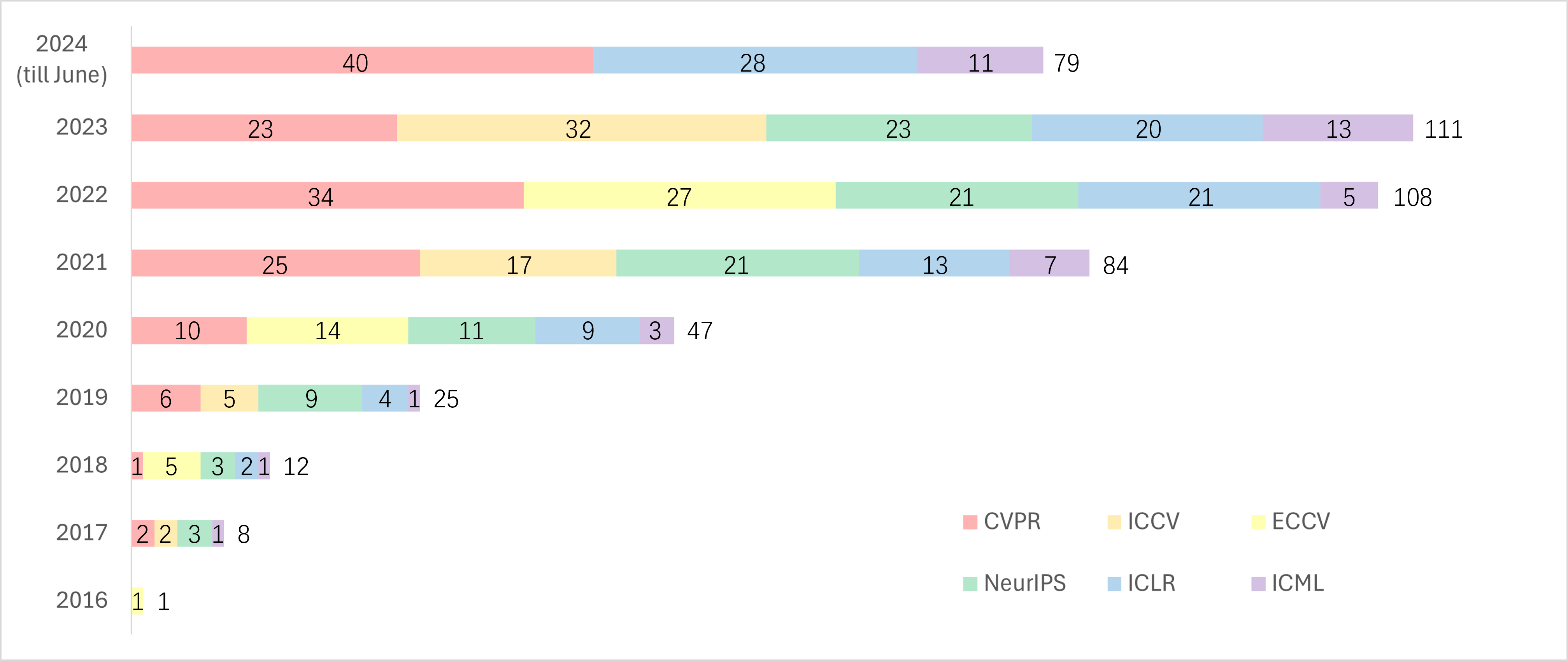}
  \caption{General trend of the number of papers on continual learning in computer vision published in top-ranked conferences during the past eight years. The plot shows consistent growth in recent literature. 
  }
  \label{fig:trend}
\end{figure*}

With the widespread popularity of cameras and mobile phones, a large number of new images and videos are captured and shared every day. This has given birth to new requirements, especially in the computer vision area, for models to learn and update themselves sequentially and continuously during its inference, since retraining a model from scratch to adapt to the daily newly generated data is time-consuming and extremely inefficient.

Considering the different structures of neural networks and human brains, neural network training is not easily transformed from its original batch learning mode to the new continual learning mode. In particular, there exist two main problems. Firstly, learning from data with multiple categories in sequential order can easily lead to the problem of catastrophic forgetting \cite{MCCLOSKEY1989109,FRENCH1999128}. This means the performance of the model on previously learned categories often decreases sharply after the model parameters are updated from the data of the new category. Secondly, when learning from new data of the same category in sequential order, this can also lead to the problem of concept drift \cite{schlimmer1986incremental,widmer1993effective,gama2014survey}, as the new data may change the data distribution of this category in unforeseen ways \cite{royer2015classifier}. Hence, the overall task of continual learning is to solve the stability-plasticity dilemma \cite{grossberg2007consciousness,mermillod2013stability}, which requires the neural networks to prevent forgetting previously learned knowledge, while maintaining the capability of learning new knowledge.

In recent years, an increasing number of continual learning methods have been proposed in various subareas of computer vision, as shown in Figure \ref{fig:trend}. Additionally, several competitions \cite{lomonaco2020cvpr, 2ndclvisioncvprworkshop, pellegrini20223rd, 4thclvisioncvprworkshop,5thclvisioncvprworkshop} related to continual learning in computer vision have been held from 2020 to 2024. Hence, in this paper, we present an overview of the recent advances of continual learning in computer vision.

Compared to existing continual learning overviews \cite{mai2021online, delange2021continual, masana2020class} focusing on image classification, we find that continual learning has been recently applied in various other computer vision subareas. Hence, in this overview, besides image classification, we also summarize the application of continual learning methods in various other subareas such as semantic segmentation, image generation, etc. 
Besides, compared with other previous continual learning surveys \cite{belouadah2020comprehensive, parisi2019continual,zhou2023deep,tian2024survey,wang2024comprehensive}, especially considering the consistent rapid development of the continual learning techniques in computer vision, in this review, we make a comprehensive summary that also focuses on the more recent and up-to-date continual learning methods.

We summarize the main contributions of this overview as follows. (1) A systematic review of the recent progress of continual learning in computer vision is provided. (2) Various continual learning techniques that are used in different computer vision tasks are introduced, including regularization, knowledge distillation, memory-based, generative replay, and parameter isolation. 
(3) The subareas in computer vision, where continual learning is potentially helpful yet still not well investigated, are discussed. 

The remainder of this paper is organized as follows. Section 2 gives the definition of continual learning.
Section 3 presents the commonly used evaluation metrics in this area. Section 4 discusses various categories of continual learning methods and their applications in computer vision. 
Section 5 conducts performance comparisons over numerous computer vision tasks.
The subareas of computer vision where continual learning has not been well exploited are then discussed in section 6. 
Finally, section 7 concludes the paper.

\section{Continual Learning: Problem Definition}

In this section, we introduce the typical formalization of continual learning, following the recent works in this area \cite{lesort2020continual,mai2021online,aljundi2019online}. We denote the initial model before continual learning as $M_0$, the number of classes of $M_0$ as $N_0$, and a potentially infinite data stream of tasks as $\mathcal{T} = \{T_1, T_2, ..., T_n, ...\}$, where $T_n = \{(X_n^i, Y_n^i)\, |\, i = 1, 2, ..., N_n\}$. $X_n^i$ and $Y_n^i$ are the sets of data instances and their corresponding labels for the $i^{th}$ class in task $T_n$, and $N_n$ is the total number of classes covered after learning $T_n$. We further denote the model after learning $T_n$ as $M_n$. Then a typical continual learning process can be defined as $\mathcal{P}$ = \{$P_1$, $P_2$, ..., $P_n$, ...\}, where $P_n : \langle M_{n-1}, T_n\rangle \longrightarrow M_n$. Note that besides a subset of data instances and labels stored in memory based methods, the data instances and labels previous to $T_n$ are generally inaccessible during the process of $P_n$. Then the general objective of the continual learning method is to ensure every model $M_n$ learned in process $P_n$ achieves good performance on the new task without affecting the performance of the previous tasks.

As mentioned by Van de Ven and Tolias \cite{van2019three}, continual learning can be further divided into three different scenarios: (i) task-, (ii) domain-, and (iii) class-incremental continual learning. Task-incremental continual learning generally requires the task ID to be given during inference. Domain-incremental continual learning aims to distinguish classes inside each task instead of distinguishing different tasks and does not require the task ID during inference. Class-incremental continual learning aims to distinguish classes both inside and among tasks, without requiring the task ID during inference. For example, suppose we want to distinguish hand-written numbers from 1 to 4, and we divide this into two tasks $T_1$ and $T_2$, where $T_1$ contains 1 and 2, and $T_2$ contains 3 and 4. For inference, with a given hand-written number from 1 to 4, task-incremental continual learning requires the corresponding task ID (either $T_1$ or $T_2$) and can distinguish between 1 and 2 in $T_1$, and between 3 and 4 in $T_2$. Meanwhile, domain-incremental continual learning does not require a task ID, but can only distinguish between odd and even numbers. In contrast, class-incremental continual learning can distinguish among the numbers 1 to 4 without requiring the task ID. Hence, in general, task-incremental and domain-incremental continual learning can be regarded as simplified versions of class-incremental continual learning. Thus, in this overview, we mainly focus on continual learning methods from the perspective of various categories of techniques, rather than their usage in specific scenarios.

\begin{figure*}
  \includegraphics[width=\textwidth]{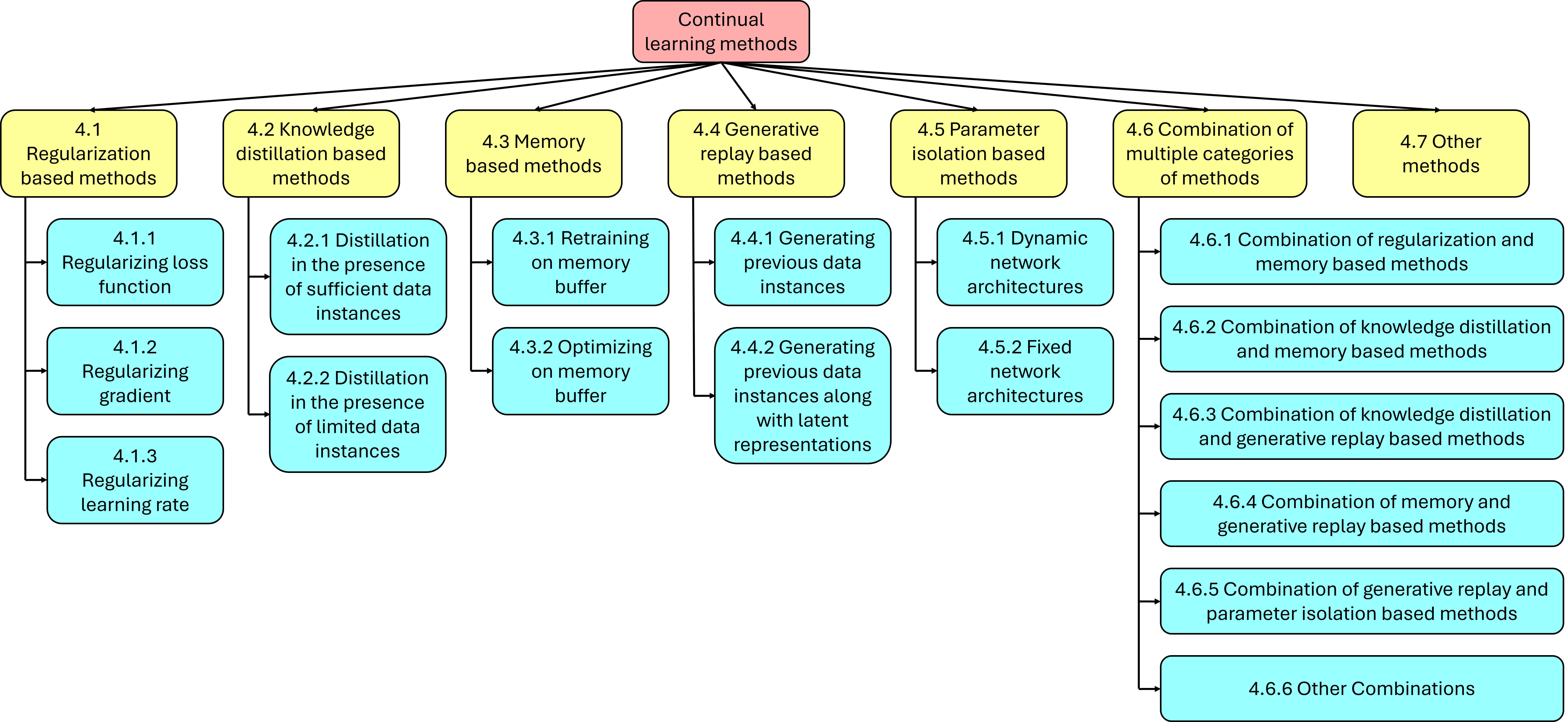}
  \caption{A taxonomy of continual learning methods.}
  \label{fig:general}
\end{figure*}

\section{Evaluation Metrics}

In the evaluation of different continual learning methods, it is important to measure both their performance on continuously learning new tasks and how much knowledge from the previous tasks has been forgotten. There are several measurement metrics \cite{lopez2017gradient,chaudhry2018riemannian} that have been popularly used in continual learning methods. In the following, we define $a_{q,p}$ as the accuracy on the test set of the $p^{th}$ task after continual learning of $q$ tasks, where $p \leqslant q$. 

\noindent\textbf{Average accuracy ($A_q$)} measures the performance of the continual learning method after a total of $q$ tasks have been learned, which can be formulated as:

\begin{equation}
\label{eq:1}
A_q = \frac{1}{q} \sum_{p=1}^{q} a_{q,p}
\end{equation}

\noindent\textbf{Average forgetting ($F_q$)} measures how much knowledge has been forgotten across the first $q-1$ tasks, 
\begin{equation}
\label{eq:3}
F_q = \frac{1}{q-1} \sum_{p=1}^{q-1} f^q_p,
\end{equation}
where knowledge forgetting $f^q_p$ of a task $p$ is defined as the difference between the maximal obtained knowledge during the continual learning process and the knowledge remaining after $q$ tasks have been learned. It can be calculated as: 
\begin{equation}
\label{eq:2}
f^q_p = \max_{o \in \{1, ..., q-1\}} a_{o, p} - a_{q, p}, \,\forall p < q.
\end{equation}

\noindent\textbf{Intransigence ($I_q$)} measures how much continual learning prevents a model from learning a new task compared to typical batch learning:

\begin{equation}
I_q = a^*_q - a_{q,q},
\end{equation}
where $a^*_q$ denotes the accuracy on the test set of the $q^{th}$ task when batch learning is used for the $q$ tasks.

\noindent\textbf{Backward transfer ($BWT_q$)} measures how much the continual learning on the $q^{th}$ task influences the performance of the previously learned tasks. It can be defined as:

\begin{equation}
BWT_q = \frac{1}{q-1} \sum_{p=1}^{q-1} (a_{q, p} - a_{p, p})
\end{equation}

\noindent\textbf{Forward transfer ($FWT_q$)} measures how much the continual learning on the $q^{th}$ task potentially influences the performance of the future tasks. Following the notation of \cite{lopez2017gradient}, we define $b_p$ as the accuracy on the test set of the $p^{th}$ task at random initialization. Then, we define forward transfer ($FWT_q$) as

\begin{equation}
FWT_q = \frac{1}{q-1} \sum_{p=2}^{q} (a_{p-1, p} - b_p)
\end{equation}

\section{Continual Learning Methods}

Recently, a large number of continual learning methods have been proposed for various computer vision tasks, such as image classification, object detection, semantic segmentation, and image generation. Generally, these methods can be divided into different categories, including regularization based, knowledge distillation based, memory based, generative replay based, and parameter isolation based methods.
Below we review these methods in detail. 
We also discuss the works that take advantage of the complementary strengths of multiple categories of methods to improve performance. 

\subsection{Regularization Based Methods}
Regularization based methods \cite{kirkpatrick2017overcoming, zenke2017continual, he2018overcoming, ebrahimi2019uncertainty} generally impose restrictions on the update process of various model parameters and hyperparameters in order to consolidate previously learned knowledge while learning new tasks to mitigate catastrophic forgetting in continual learning. This can be achieved through a variety of schemes, which are introduced below.

\subsubsection{Regularizing Loss Function}

As the name suggests, the most typical scheme used by regularization based methods is to consolidate previously learned knowledge by regularizing the loss function. 

\begin{figure}
  \includegraphics[width=\linewidth]{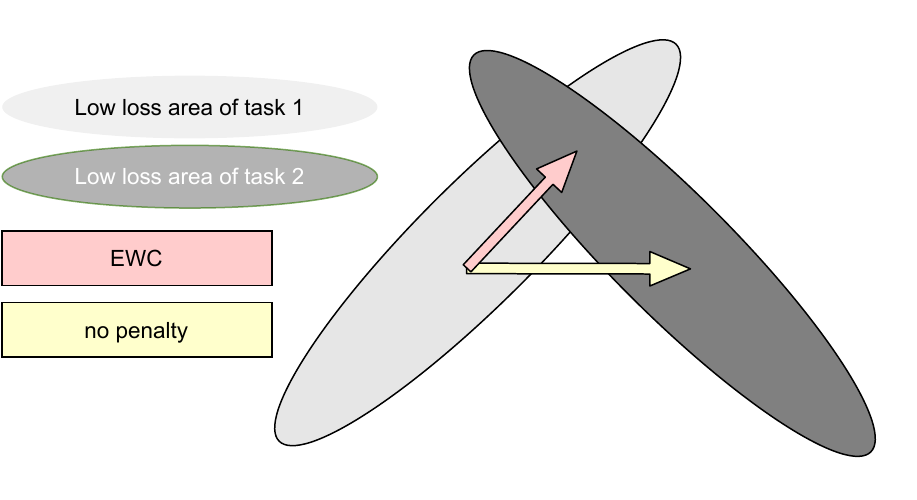}
  \caption{Illustration of EWC \cite{kirkpatrick2017overcoming}, which bounds the model parameters to update to the common low loss area among tasks, instead of the low loss area of only the new task.}
  \label{fig:regularization}
\end{figure}

In image classification, a typical method called Elastic Weight Consolidation (EWC) was first proposed by Kirkpatrick et al. \cite{kirkpatrick2017overcoming}. EWC injects a new quadratic penalty term into the loss function to restrict the model from modifying the weights that are important for the previously learned tasks. The importance of weights is calculated via the diagonal of the Fisher information matrix. As shown in Figure \ref{fig:regularization}, by adding the new penalty term to the loss function, EWC bounds the model parameters to update to the common low loss area among tasks, instead of the low loss area of the new task only, thus helping to alleviate the catastrophic forgetting problem. 

However, the assumption of EWC \cite{kirkpatrick2017overcoming}, that the Fisher information matrix is diagonal, is almost never true. To address this issue, Liu et al. \cite{liu2018rotate} proposed to approximately diagonalize the Fisher information matrix by rotating the parameter space of the model, leaving the forward output unchanged. Besides, EWC \cite{kirkpatrick2017overcoming} also requires a quadratic penalty term to be added for each learned task, and hence has a linearly increasing computational cost. To handle this issue, Schwarz et al. \cite{schwarz2018progress} proposed an online version of EWC to extend EWC to a memory-efficient version by only focusing on a single penalty term on the most recent task. Around the same time, Chaudhry et al. \cite{chaudhry2018riemannian} also proposed an efficient alternative to EWC called EWC++ which maintains a single Fisher information matrix for all the previously learned tasks and updates the matrix using moving average \cite{martens2015optimizing}. Besides the above, Loo et al. \cite{loo2020generalized} proposed Generalized Variational Continual Learning which unifies both Laplacian approximation and variational approximation. The authors also proposed a task-specific layer to further mitigate the over-pruning problem in variational continual learning. 
Lee et al. \cite{lee2017overcoming} merged Gaussian posteriors of models trained on old and new tasks respectively by either mean-incremental-moment-matching which simply averages the parameters of two models or mode-incremental-moment-matching which utilizes Laplacian approximation to calculate a mode of the mixture of two Gaussian posteriors. Ritter et al. \cite{ritter2018online} updated a quadratic penalty term in the loss function for every task by the block-diagonal Kronecker factored approximation of the Hessian matrix for considering the intra-layer parameter interaction simultaneously. 
Lee et al. \cite{lee2020continual} pointed out that the current usage of the Hessian matrix approximation as the curvature of the quadratic penalty function is not effective for networks containing batch normalization layers. To resolve this issue, a Hessian approximation method, which considers the effects of batch normalization layers was proposed. 

Aside from EWC \cite{kirkpatrick2017overcoming} and its extensions \cite{liu2018rotate,schwarz2018progress} that generally calculate the importance of weights by approximating curvature, Zenke et al. \cite{zenke2017continual} proposed Synaptic Intelligence (SI) to measure the weights' importance with the help of synapses. They pointed out that one-dimensional weights are too simple to preserve knowledge. Hence, they proposed three-dimensional synapses which can preserve much more knowledge and prevent important synapses from changing to preserve important previously learned knowledge. A modified version of SI, which measures the importance of weights using the distance in the Riemannian manifold instead of the Euclidean distance, was also proposed in \cite{chaudhry2018riemannian} to effectively encode the information about all the previous tasks. Park et al.  \cite{park2019continual} pointed out that SI \cite{zenke2017continual} may underestimate the loss since it assumes the loss functions are symmetric which is often incorrect. Hence, they proposed Asymmetric Loss Approximation with Single-Side Overestimation (ALASSO), which considers the loss functions of the previous tasks as the observed loss functions. Using the quadratic approximation of these observed loss functions, ALASSO derives the loss function required for the new task by overestimating the unobserved part of the previous loss functions. 

In addition to EWC \cite{kirkpatrick2017overcoming} and SI \cite{zenke2017continual}, various other methods \cite{aljundi2018memory, ren2018incremental} have been proposed to preserve important parameters from other perspectives. Aljundi et al. \cite{aljundi2018memory} proposed a method called Memory Aware Synapses (MAS). MAS calculates the importance of weights with a model of Hebbian learning in biological systems, which relies on the sensitivity of the output function and can hence be utilized in an unsupervised manner. Ren et al. \cite{ren2018incremental} focused on incremental few-shot learning. They proposed a method called Attention Attractor Network which adds an additional penalty term, adapted from attractor networks \cite{zemel2001localist}, to the loss function. Cha et al. \cite{cha2020cpr} pointed out that as the number of learned tasks increases, the common low loss area for all the learned tasks quickly becomes quite small. Hence, they proposed Classifier-Projection Regularization which adds an additional penalty term to the loss function to widen the common low loss area. Hu et al. \cite{hu2021continual} proposed Per-class Continual Learning which treats every task holistically on itself. Hence, new tasks would be less likely to modify the features of those learned tasks. This is achieved by adapting the one-class loss with holistic regularization \cite{hu2020hrn} in continual learning scenarios.

Further to methods that add a single penalty term to the loss function, some methods \cite{ahn2019uncertainty, jung2020continual} handle the stability-plasticity dilemma by directly adding one penalty term for stability and one for plasticity. 
Ahn et al. \cite{ahn2019uncertainty} integrated the idea of uncertainty into regularization by making the variance of the incoming weights of each node trainable and further added two additional penalty terms respectively for stability and plasticity. 
Jung et al. \cite{jung2020continual} selectively added two additional penalty terms to the loss function, comprising a Lasso term that controls the ability of the model to learn new knowledge, and a drifting term to prevent the model from forgetting. Volpi et al. \cite{volpi2021continual} further proposed a meta-learning strategy to consecutively learn from different visual domains. Two penalty terms are added to the loss function, comprising a recall term to mitigate catastrophic forgetting and an adapt term to ease adaptation to each new visual domain. To mimic new visual domains, the authors applied heavy image manipulations to the data instances from the current domain to generate multiple auxiliary meta-domains. Aside from image classification, they also adapted their model to perform semantic segmentation. 

Beyond image classification, some works have focused on other computer vision problems, such as domain adaptation \cite{kundu2020class}, image generation \cite{seff2017continual} and image de-raining \cite{zhou2021image}. Kundu et al. \cite{kundu2020class} combined the problem of image classification with domain adaptation and proposed a modified version of the prototypical network \cite{snell2017prototypical} to solve this proposed new problem. Seff et al. \cite{seff2017continual} adapted EWC to a class-conditional image generation task, which requires the model to generate new images conditioned on their class. Zhou et al. \cite{zhou2021image} proposed to apply a regularization based method on image de-raining. They regarded each dataset as a separate task and proposed a Parameter-Importance Guided Weights Modification approach to calculate the task parameter importance using the Hessian matrix approximation calculated by the Jacobian matrix for storage efficiency.

\subsubsection{Regularizing Gradient}

Besides regularizing loss functions, a few other methods \cite{he2018overcoming, zeng2019continual} proposed to regularize the gradient given by backpropagation to prevent the update of parameters from interfering with previously learned knowledge.

In image classification, He and Jaeger \cite{he2018overcoming} replaced the typical backpropagation with conceptor-aided backpropagation. For each layer of the network, a conceptor characterizing the subspace of the layer spanned by the neural activation appearing in the previous tasks is calculated and this conceptor is preserved during the backpropagation process. Moreover, Zeng et al. \cite{zeng2019continual} proposed an Orthogonal Weights Modification algorithm. During the training process of each new task, the modification of weights calculated during typical backpropagation is further mapped onto a subspace generated by all the previous tasks in order to maintain the performance of the previous tasks. A Context-Dependent Processing module is also included to facilitate the learning of contextual features. Similarly, Wang et al. \cite{wang2021training} pointed out that mapping the update of the model parameters for learning the new task into the null space of the previous tasks can help to mitigate catastrophic forgetting. They proposed to use Singular Value Decomposition to approximate the null space of the previous tasks. Later on, Saha et al. \cite{saha2021gradient} partitioned the gradient into Core Gradient Space and Residual Gradient Space which are orthogonal to each other. The gradient step of the new task is enforced to be orthogonal to the Core Gradient Spaces of the previous tasks. As a result, learning the new task can minimally affect the performance of the model on the previous tasks. 

More recently, as an extension of \cite{saha2021gradient},  Qiao et al. \cite{qiao2024prompt} further explored how the ``orthogonal gradients'' idea can be well-incorporated into the prompt tuning paradigm \cite{wang2022s,li2024steering}. Similarly, Xiao et al. \cite{xiao2024hebbian} got inspired from the ``orthogonal gradients'' idea as well. However, according to Xiao et al. \cite{xiao2024hebbian}, especially in the context of Spiking Neural Networks, operations such as matrix inversion and singular value decomposition that are required in \cite{saha2021gradient} can be hard to implement. Accordingly, they proposed to perform orthogonal gradient projections in a Hebbian-based manner instead. Also at a recent time, Elsayed and Mahmood \cite{elsayed2024addressing} proposed to regularize the gradient through a Utility-based Perturbed Gradient Descent mechanism. Via approximating the utility (usefulness) of the weight units and leveraging them during the gradient descent process, this mechanism can prevent those already useful weights from being overly updated, while at the same time perturbing those unuseful weights and easing their updates. 

\subsubsection{Regularizing Learning Rate}
Aside from the regularization of model parameters, a few other methods \cite{ebrahimi2019uncertainty, mirzadeh2020understanding} proposed to regularize the learning rate, which has been shown to be effective in mitigating catastrophic forgetting.

In image classification, Ebrahimi et al. \cite{ebrahimi2019uncertainty} utilized uncertainty to modify the learning rate of the model parameters based on their importance for the previous tasks. Mirzadeh et al. \cite{mirzadeh2020understanding} pointed out that the properties of the local minima of each task have an important role in preventing forgetting. Hence, they proposed to tune the learning rate and batch size to indirectly control the geometry of the local minima of different tasks.

In image semantic segmentation, Ozgun et al. \cite{ozgun2020importance} proposed to prevent the model from losing knowledge by restricting the adaptation of important model parameters with learning rate regularization. More precisely, the learning rate of important model parameters is reduced, while the learning rate of non-important parameters is kept the same.

Apart from the above-mentioned schemes, i.e., regularizing the loss function, gradient, and learning rate, Kapoor et al. \cite{kapoor2021variational} performed regularization from another perspective. They proposed Variational Auto-Regressive Gaussian Processes, which use sparse inducing point approximation to better approximate the Gaussian posterior, resulting in a lower bound objective for regularization.

\subsection{Knowledge Distillation Based Methods}

Knowledge distillation based methods \cite{li2017learning, dhar2019learning} incorporate the idea of knowledge distillation into continual learning by distilling knowledge from the model trained on the previous tasks to the model trained on the new task in order to consolidate previously learned knowledge. This can be achieved through a variety of schemes, which are introduced below.

\subsubsection{Distillation in the Presence of Sufficient Data Instances}

The majority of knowledge distillation based methods \cite{li2017learning, dhar2019learning} were designed for continual learning from a stream of sufficient data instances for each new task. 

\begin{figure}
  \includegraphics[width=\linewidth]{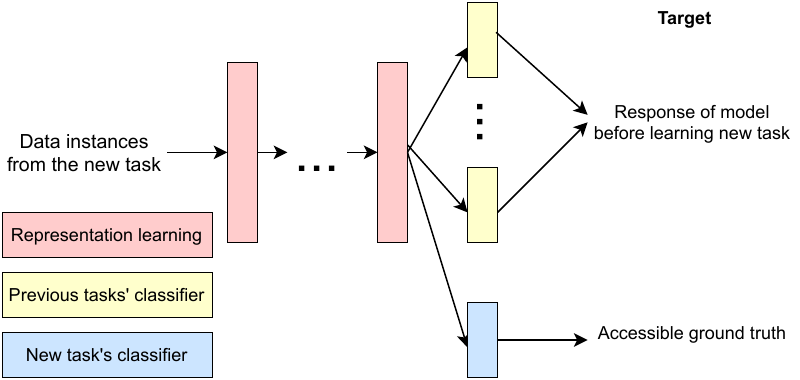}
  \caption{Illustration of LwF \cite{li2017learning}, which stores a copy of the previous model's parameters before learning the new task, and uses the response of that copied model on the data instances from the new task as the target for the previous tasks' classifiers during the learning of the new task, while using the accessible ground truth as the target for the new task classifier.}
  \label{fig:kd}
\end{figure}

In image classification, Li and Hoiem \cite{li2017learning} proposed a typical method called Learning without Forgetting (LwF). As shown in Figure \ref{fig:kd}, LwF stores a copy of the previous model parameters before learning the new task. It then uses the response of that copied model on the data instances from the new task as the target for the classifiers of the previous tasks. The accessible ground truth is used as the target for the new task classifier. However, Rannen et al. \cite{rannen2017encoder} pointed out that LwF would not result in good performance if the data distributions between different tasks are quite diverse. Hence, they further trained an autoencoder for each task to learn the most important features corresponding to the task, and used it to preserve knowledge. 

As time passed, Dhar et al. \cite{dhar2019learning} proposed Learning without Memorizing, which adds an extra term i.e., attention distillation loss, to the knowledge distillation loss. The attention distillation loss penalizes changes in the attention maps of classifiers and helps to retain previously learned knowledge. Later on, Fini et al. \cite{fini2020online} proposed a two-stage method called Batch-Level Distillation. In the first stage of learning, only the data instances of the new task are used to minimize the classification loss over the new task classifier. In the second stage of learning, both the knowledge distillation and the learning of the new task are carried out simultaneously.
Moreover, Douillard et al. \cite{douillard2020podnet} regarded continual learning as representation learning and proposed a distillation loss called Pooled Outputs Distillation which constrains the update of the learned representation from both the final output and the intermediate layers. Kurmi et al. \cite{kurmi2021not} utilized the prediction uncertainty of the model on the previous tasks to mitigate catastrophic forgetting. Aleatoric uncertainty and self-attention are involved in the proposed distillation loss. Simon et al. \cite{simon2021learning} proposed to conduct knowledge distillation on the low-dimensional manifolds between the model outputs of previous and new tasks, which was shown to better mitigate catastrophic forgetting. Besides, Hu et al. \cite{hu2021distilling} found that the knowledge distillation based approaches do not have a consistent causal effect compared to end-to-end feature learning. Hence, they proposed to also distill the colliding effect as a complementary method to preserve the causal effect.

In addition to mitigating catastrophic forgetting, a few other methods \cite{hou2019learning, zhao2020maintaining} were also proposed to solve the data imbalance problem or the concept drift problem at the same time.
Hou et al. \cite{hou2019learning} systematically investigated the problem of data imbalance between the previous and new data instances and proposed a new knowledge distillation based framework that treats both previous and new tasks uniformly to mitigate the effects of data imbalance. A cosine normalization, a less-forget constraint that acts as a feature level distillation loss, and an inter-class separation component are incorporated to effectively address data imbalance from different aspects. Zhao et al. \cite{zhao2020maintaining} pointed out that the parameters of the last fully connected layer are highly biased in continual learning. They thus proposed a Weight Aligning method to correct the bias towards new tasks. He et al. \cite{he2020incremental} pointed out that concept drift is much less explored in previous continual learning studies compared to catastrophic forgetting. Hence, the authors proposed both a new cross distillation loss to handle catastrophic forgetting and an exemplar updating rule to handle concept drift.

A few approaches \cite{ke2020continual,lee2021sharing} have been proposed to only transfer relevant knowledge. Ke et al. \cite{ke2020continual} proposed Continual learning with forgetting Avoidance and knowledge Transfer (CAT). When a new task comes, CAT automatically distinguishes the previous tasks into similar tasks and dissimilar tasks. After that, knowledge is transferred from those similar tasks to facilitate the learning of the new task, and task masks are applied to those dissimilar tasks to avoid them being forgotten. Lee et al. \cite{lee2021sharing} proposed to only transfer knowledge at a selected subset of layers. The selection is made based on an Expectation-Maximization method.

Besides image classification, several recent works \cite{michieli2019incremental,cermelli2020modeling,douillard2021plop,michieli2021continual,xiao2023endpoints,gong2024continual,wang2024incremental} focused on image semantic segmentation. Michieli et al. \cite{michieli2019incremental} proposed several approaches to distill the knowledge of the model learned on the previous tasks, whilst updating the current model to learn the new ones. Cermelli et al. \cite{cermelli2020modeling} further pointed out that at each training step, as the label is only given to areas of the image corresponding to the learned classes, other background areas suffer from a semantic distribution shift, which is not considered in previous works. They then proposed both a new distillation based framework and a classifier parameter initialization strategy to handle the distribution shift. Douillard et al. \cite{douillard2021plop} handled the background semantic distribution shift problem by generating pseudo-labels of the background from the previously learned model. They also proposed a multi-scale spatial distillation, preserving both long-range and short-range spatial relationships at the feature level to mitigate catastrophic forgetting. Michieli and Zanuttigh \cite{michieli2021continual} proposed several strategies in the latent space complementary to knowledge distillation including prototype matching, contrastive learning, and feature sparsity. Xiao et al. \cite{xiao2023endpoints} incorporated the idea of knowledge distillation together with the idea of model ensemble. Specifically, they proposed to first distill the knowledge of the previously learned model during the learning process to learn a temporary version of the current model. After that, they fused the weights of the previously learned model and the weights of the temporary version of the current model to get the final version of the current model. 

More recently, Gong et al. \cite{gong2024continual} observed that compared to per-pixel segmentation models, query-based model architectures such as Mask2Former \cite{cheng2022masked} can provide continual semantic segmenters with stronger forward knowledge transferring ability, while at the same time let them less suffer from catastrophic forgetting. Considering this, Gong et al. \cite{gong2024continual} proposed a query-based framework, while further incorporating the framework with both objectness distillation and class distillation strategies. Around the same time, Wang et al. \cite{wang2024incremental} proposed another continual semantic segmentation method, while particularly focusing their method on the segmentation of nuclei from histopathological images. Specifically, in their proposed method, they equipped a compatibility-inspired distillation scheme to better align between the model learned from the previous tasks and the model learned from the new task. Moreover, they also incorporated a future-class awareness mechanism to improve the model's forward knowledge transferring ability and better prepare the model for future tasks.

Additionally, the application of continual learning in the person re-identification task has also attracted research attention in the past few years \cite{pu2021lifelong,sun2022patch,cui2024learning}. Among them, Pu et al. \cite{pu2021lifelong} proposed to apply knowledge distillation on person re-identification for the first time. Specifically, they pointed out that in human cognitive science, the brain was found to focus more on stabilization during knowledge representation and more on plasticity during knowledge operation. Hence, an Adaptive Knowledge Accumulation method was proposed as an alternative to the typical knowledge distillation method. This method uses a knowledge graph as knowledge representation and a graph convolution as knowledge operation, and further involves a plasticity-stability loss to mitigate catastrophic forgetting.
More recently, Cui et al. \cite{cui2024learning} further particularly focused on the setting of re-indexing free lifelong person re-identification. Specifically, inspired by compatible training \cite{pan2023boundary,ramanujan2022forward,seo2023metric}, they first proposed the Continual Compatible Representation which essentially serves as an extension of compatible training in the continual learning scenario with large domain gaps between tasks. Moreover, Cui et al. \cite{cui2024learning} also further incorporated their method with two distillation modules (i.e., the balanced compatible distillation module and the balanced anti-forgetting distillation module) to further strengthen their method.

Aside from image classification, image semantic segmentation, and person re-identification, some works have focused on other computer vision problems, such as object detection \cite{shmelkov2017incremental,feng2022overcoming}, conditional image generation \cite{wu2018memory,zhai2019lifelong}, and image and video captioning \cite{nguyen2019contcap}.
Shmelkov et al. \cite{shmelkov2017incremental} adapted the knowledge distillation method for object detection from images. This method distills knowledge from both the unnormalized logits and the bounding box regression output of the previous model. 
After Shmelkov et al. \cite{shmelkov2017incremental}, Feng et al. \cite{feng2022overcoming} further found that, in the context of continual object detection, it is important to assign knowledge from different responses with different importance levels. They then proposed a method named Elastic Response Distillation to perform importance level assignment.
Wu et al. \cite{wu2018memory} adapted the knowledge distillation method for class-conditioned image generation. They proposed an aligned distillation method to align the generated images of the auxiliary generator with those generated by the current generator, given the same label and latent feature as inputs. Zhai et al. \cite{zhai2019lifelong} also applied knowledge distillation to conditional image generation. They proposed a Lifelong Generative Adversarial Network (Lifelong GAN) to adapt knowledge distillation toward class-conditioned image generation task that requires no image storage. As a result, this method can also be used for other conditional image generation tasks such as image-conditioned generation, where a reference image instead of a class label is given and there may exist no related image in the previous tasks. Nguyen et al. \cite{nguyen2019contcap} applied knowledge distillation to both image and video captioning. For image captioning, they proposed to adapt pseudo-label methods. Besides, they also merged knowledge distillation of intermediate features and proposed to partly freeze the model to transfer knowledge smoothly while maintaining the capability to learn new knowledge.

In contrast to the above-mentioned methods, Lee et al. \cite{lee2019overcoming} pointed out that increasing the number of data instances for training each task can help mitigate catastrophic forgetting. They thus defined a new problem setup where unlabelled data instances are used together with labeled data instances during continual learning. A confidence-based sampling method was proposed to select unlabelled data instances similar to the previous tasks. They further proposed a global distillation method to distill the knowledge from all the previous tasks together instead of focusing on each previous task separately.

\subsubsection{Distillation in the Presence of Limited Data Instances}

Aside from knowledge distillation based continual learning with sufficient data instances per each new task, several recent works \cite{cheraghian2021semantic,yoon2020xtarnet, perez2020incremental} have focused on few-shot continual learning where only a few data instances are given for each new task. 

In image classification, Liu et al. \cite{liu2020incremental} proposed a method named Indirect Discriminant Alignment (IDA). IDA does not align the classifier of the new task towards all the previous tasks during distillation but carries out alignment to a subset of anchor tasks. Hence, the model is much more flexible towards learning new tasks. 
Cheraghian et al. \cite{cheraghian2021semantic} made use of word embeddings when only a few data instances are available. Semantic information from word embeddings is used to identify the shared semantics between the learned tasks and the new task. These shared semantics are then used to facilitate the learning of the new task and preserve previously learned knowledge. 
Yoon et al. \cite{yoon2020xtarnet} proposed a few-shot continual learning framework containing a base model and different meta-learnable modules. When learning a new task, novel features, which are extracted by the meta-learnable feature extractor module, are combined with the base features to produce a task-adaptive representation. The combination process is controlled by another meta-learnable module. The task-adaptive representation helps the base model to quickly adapt to the new task. 
More recently, Park et al. \cite{park2024pre} further explored the usage of the pre-trained vision and language models in few-shot continual learning. Specifically, they proposed a semantic knowledge distillation loss to facilitate the learning process of every new visual task with knowledge from the language space, while also introducing a pre-trained knowledge tuning process for the base training process of few-shot continual learning. In object detection, Perez et al. \cite{perez2020incremental} proposed OpeNended Centre nEt (ONCE) which adapts CentreNet \cite{zhou2019objects} detector towards the continual learning scenario where new tasks are registered with the help of meta learning.

Besides the above-mentioned schemes, i.e., distillation in the presence of sufficient data instances and distillation in the presence of limited data instances, Yoon et al. \cite{yoon2021federated} proposed a new problem setup called federated continual learning, which allows several clients to carry out continual learning each by itself through an independent stream of data instances that is inaccessible to other clients. To solve this problem, they proposed to separate parameters into global and task-specific parameters, whereby each client can obtain a weighted combination of the task-specific parameters of the other clients for knowledge transfer.

\subsection{Memory Based Methods}

Memory based methods \cite{rebuffi2017icarl, chaudhry2019tiny, lopez2017gradient} generally have a memory buffer to store data instances and/or various other information related to the previous tasks, which are replayed during the learning of new tasks, in order to consolidate previously learned knowledge to mitigate catastrophic forgetting. This can be achieved through a variety of schemes, which we introduce below.

\subsubsection{Retraining on Memory Buffer}
Most memory based methods \cite{rebuffi2017icarl, chaudhry2019tiny} preserve previously learned knowledge by retraining on the data instances stored in the memory buffer.

In image classification, Rebuff et al. \cite{rebuffi2017icarl} first proposed a typical method called incremental Classifier and Representation Learning (iCARL). During representation learning, iCARL utilizes both the stored data instances and the instances from the new task for training. During classification, iCARL adopts a nearest-mean-of-exemplars classification strategy to assign the label of the given image to the class with the most similar prototype. The distance between data instances in the latent feature space is used to update the memory buffer. The original iCARL method requires all data from the new task to be trained together. To address this limitation and enable the new instances from a single task to come at different time steps, Chaudhry et al. \cite{chaudhry2019tiny} proposed Experience Replay (ER), which uses reservoir sampling \cite{vitter1985random} to randomly sample a certain number of data instances from a data stream of unknown length, and store them in the memory buffer. 

However, reservoir sampling \cite{vitter1985random} works well if each of the tasks has a similar number of instances, and it could lose information from the tasks that have significantly fewer instances than the others. Thus, several other sampling algorithms \cite{aljundi2019gradient, liu2020mnemonics} were proposed to address this issue. Aljundi et al. \cite{aljundi2019gradient} regarded the selection of stored data instances as a constraint selection problem and sampled data instances that can minimize the solid angle formed by their corresponding constraints. To reduce the computational cost, they further proposed a greedy version of the original selection algorithm. Liu et al. \cite{liu2020mnemonics} trained exemplars using image-size parameters to store the most representative data instances from the previous tasks. Kim et al. \cite{kim2020imbalanced} proposed a partitioning reservoir sampling as a modified version of reservoir sampling to address the data imbalance problem.
Chrysakis and Moens \cite{chrysakis2020online} proposed Class-Balancing Reservoir Sampling (CBRS) as an alternative to reservoir sampling. CBRS is a two-phase sampling technique. During the first phase, all new data instances are stored in the memory buffer as long as the memory is not filled. After the memory buffer is filled, the second phase is activated to select which stored data instance needs to be replaced with the new data instance. Specifically, the new instance is replaced with a stored data instance from the same class if it belongs to a class that dominates the memory buffer at the time or at some previous time stamps. Otherwise, it is replaced with a stored data instance from the class that dominates the memory buffer at the time. 
Borsos et al. \cite{borsos2020coresets} proposed an alternative reservoir sampling method named coresets, which stores a weighted subset of data instances from each previous task in the memory buffer through a bilevel optimization with cardinality constraints. 

More than just focusing on sampling, more recently, Buzzega et al. \cite{buzzega2021rethinking} applied a total of five different tricks, including Independent Buffer Augmentation, Bias Control, Exponential Learning Rate Decay, Balanced Reservoir Sampling, and Loss-Aware Reservoir Sampling, on ER \cite{chaudhry2019tiny} and other methods to show their applicability. Moreover, Bang et al. \cite{bang2021rainbow} proposed Rainbow Memory to select previous data instances, particularly in scenarios where different tasks can share the same classes. It uses classification uncertainty and data augmentation to improve the diversity of data instances. Besides, similar to \cite{joseph2021towards}, Lee et al. \cite{lee2024continual} also proposed to store a balanced subset of previous data instances. Yet, inspired by \cite{kirichenko2023last}, Lee et al. \cite{lee2024continual} proposed to only retrain the last layer instead of the whole model leveraging these stored data instances. Around the same time yet from another new perspective, Wu et al. \cite{wu2024mitigating} proposed to incorporate Pareto optimization into memory-based continual learning, which enabled the interdependence of previously learned tasks to be also considered during the learning process of the new task.

There are also a few other methods \cite{aljundi2019online, shim2021online} that focus on selecting data instances from the memory buffer for retraining. Aljundi et al. \cite{aljundi2019online} proposed a method called Maximally Interfered Retrieval to select a subset of stored data instances, which suffer from an increase in loss if the model parameters are updated based on new data instances. Shim et al. \cite{shim2021online} proposed an Adversarial Shapley value scoring method, which selects the previous data instances that can mostly preserve their decision boundaries during the training of the new task.

Besides methods that store previous data instances in a single buffer, there are also a few other methods \cite{wu2019large, pham2020contextual} that divide the memory buffer into several parts. Wu et al. \cite{wu2019large} pointed out that current methods that perform well on small datasets with only a few tasks cannot maintain their performance on large datasets with thousands of tasks. Hence, they proposed the Bias Correction (BiC) method focusing on continual learning from large datasets. Noticing that a strong bias towards every new task actually exists in the classification layer, BiC specifically splits a validation set from the combination of the previous and new data instances and adds a linear bias correction layer after the classification layer to measure and then correct the bias using the validation set. Pham et al. \cite{pham2020contextual} separated the memory into episodic memory and semantic memory. The episodic memory is used for retraining and the semantic memory is for training a controller that modifies the parameter of the base model for each task.

Some other methods \cite{belouadah2019il2m, chaudhry2020using} have also been proposed to store other information together with previous data instances. 
Belouadah and Popescu \cite{belouadah2019il2m} pointed out that initial statistics of the previous tasks could help to rectify the prediction scores of the previous tasks. They thus proposed Incremental Learning with Dual Memory which has a second memory to store statistics of the previous tasks. These statistics are then used as complementary information to handle the data imbalance problem. In their subsequent work \cite{belouadah2020scail}, the initial classifier of each task is stored in separate memory. After the training of each new task, the classifier of each previous task is replaced by a scaled version of its initial stored classifier with the help of aggregate statistics. Another similar method \cite{belouadah2020initial} was also proposed to standardize the stored initial classifier of each task, resulting in a fair and balanced classification among different tasks. Chaudhry et al. \cite{chaudhry2020using} proposed Hindsight Anchor Learning to store an anchor per task in addition to data instances. The anchor is selected by maximizing a forgetting loss term. As such, the anchor can be regarded as the easiest-forgetting point of each task. By keeping these anchors to predict correctly, the model is expected to mitigate catastrophic forgetting. Ebrahimi et al. \cite{ebrahimi2020remembering} stored both data instances and their model visual explanations (i.e., saliency maps) in the memory buffer, and encouraged the model to remember the visual explanations for the predictions.

A few other methods were also proposed to store either compressed data instances \cite{hayes2020remind} or activation volumes \cite{pellegrini2020latent}. Hayes et al. \cite{hayes2020remind} proposed to store a compressed representation of previous data instances obtained by performing product quantization \cite{jegou2010product}. Aside from image classification, Hayes et al. have also applied their method to Visual Question Answering (VQA) to show its generalizability. Pellegrini et al. \cite{pellegrini2020latent} stored activation volumes of previous data instances obtained from the intermediate layers, leading to much faster computation speed.

Aside from image classification, a few works have focused on other computer vision problems, such as image semantic segmentation \cite{tasar2019incremental,zhu2023continual}, object detection \cite{joseph2021towards}, and analogical reasoning \cite{hayes2021selective}. With respect to the image semantic segmentation task, Tasar et al. \cite{tasar2019incremental} particularly focused on satellite imagery, by considering segmentation as a multi-task learning problem where each task represents a binary classification. They stored patches from previously learned images and trained them together with data instances from new tasks. 
Besides Tasar et al. \cite{tasar2019incremental}, Zhu et al. \cite{zhu2023continual} also focused on applying continual learning in the task of image semantic segmentation. Specifically, they designed a reinforcement-learning-based strategy to enable suitable data instances to be selected to refill the memory buffer at each continual learning stage. 

More recently, Joseph et al. \cite{joseph2021towards} applied the memory based method on open-world object detection and stored a balanced subset of previous data instances to fine-tune the model after learning every new task. 
Hayes and Kanan \cite{hayes2021selective} proposed an analogical reasoning method, where Raven's Progressive Matrices \cite{zhang2019raven} are commonly used as the measurement method. They used selective memory retrieval, which is shown to be more effective compared to random retrieval in continual analogical reasoning. 
Moreover, Ye and Bors \cite{ye2024online} further explored continual learning in image generation, and they proposed a new memory management method DCM that can be used in both supervised and unsupervised continual learning scenarios. Specifically, to perform memory management in high quality, DCM is proposed to be incorporated with various different designs w.r.t. data selection, memory expansion, and memory pruning.
Around the same time, Duan et al. \cite{duan2024towards} focused on applying continual learning in the image compression task. Specifically, during the training process of a new task, inspired by \cite{rebuffi2017icarl}, Duan et al. \cite{duan2024towards} proposed to perform training leveraging both the data instances from the new task and those stored data instances from the previous tasks. Besides, Duan et al. \cite{duan2024towards} also drew inspiration from the hierarchical residual coding architecture \cite{duan2023qarv,feng2023nvtc,zhu2022unified} and correspondingly proposed a new model structure for the continual image compressor.

\subsubsection{Optimizing on Memory Buffer}

As an alternative to the retraining scheme, which may over-fit on the stored data instances, the optimizing scheme \cite{lopez2017gradient,chaudhry2018efficient} has been proposed to only use the stored data instances to prevent the training process of the new task from interfering with previously learned knowledge but not retraining on them. 

In image classification, Lopez and Ranzato \cite{lopez2017gradient} proposed a typical method called Gradient Episodic Memory (GEM), which builds an inequality constraint to prevent the parameter update from increasing the loss of each individual previous task, approximated by the instances in the episodic memory, during the learning process of the new task. 
Chaudhry et al. \cite{chaudhry2018efficient} further proposed an efficient alternative to GEM named Averaged GEM which tries to prevent the parameter update from increasing the average episodic memory loss. 
Sodhani et al. \cite{sodhani2020toward} proposed to unify GEM and Net2Net \cite{chen2016accelerating} frameworks to enable the model to mitigate catastrophic forgetting and increase its capacity.

Aside from GEM \cite{lopez2017gradient} and its extensions \cite{chaudhry2018efficient}, other methods \cite{derakhshani2021kernel, tang2021layerwise} were also proposed to optimize the model on the stored data instances from various other perspectives. Derakhshani et al. \cite{derakhshani2021kernel} proposed Kernel Continual Learning, which uses the stored data instances to train a non-parametric classifier with kernel ridge regression. Tang et al. \cite{tang2021layerwise} separated the gradient of the stored data instances of the previous tasks into shared and task-specific parts. The gradient for the update is then enforced to be consistent with the shared part but orthogonal to the task-specific part. With this requirement, the common knowledge among the previous tasks can help the learning of the new task while the task-specific knowledge remains invariant to mitigate catastrophic forgetting.

Preserving the topology of previously learned knowledge has also recently become another interesting direction. Tao et al. \cite{tao2020few} proposed to store the topology of the feature space of the previous tasks through a neural gas network \cite{martinetz1991neural} and proposed the TOpology-Preserving knowledge InCrementer framework to preserve the neural gas topology of the previous tasks and adapt to new tasks given a few data instances. Tao et al. \cite{tao2020topology} proposed Topology-Preserving Class Incremental Learning (TPCIL) to store an Elastic Hebbian Graph (EHG) instead of previous data instances in the buffer. EHG is constructed based on competitive Hebbian learning \cite{martinetz1993competitive} and represents the topology of the feature space. Inspired by the idea from human cognitive science that forgetting is caused by breaking topology in human memory, TPCIL mitigates catastrophic forgetting by injecting a topology-preserving term into the loss function.

Besides topology information, a few other methods \cite{von2019continual,iscen2020memory} proposed to preserve various other information. Oswald et al. \cite{von2019continual} proposed to store a task embedding for each task in the buffer. Then, given the task embedding, a task-conditioned hypernetwork is trained to output the corresponding model parameters. Iscen et al. \cite{iscen2020memory} proposed to store only the feature descriptors of the previous tasks. When a new task comes, instead of co-training the new data instances with previous ones, the method conducts a feature adaptation between the stored feature descriptor and the feature descriptor of the new task. Ren et al. \cite{ren2020wandering} introduced a new problem setup called Online Contextualized Few-Shot Learning. They then proposed to store both a prototype per class and a contextual prototypical memory focusing particularly on the contextual information. Zhu et al. \cite{zhu2021prototype} proposed to store a class-representative prototype for each previous task and augment these prototypes to preserve the decision boundaries of the previous tasks. Self-supervised learning was also proposed to generalize features from the previous tasks, facilitating the learning of the new task. 

In addition, Joseph and Balasubramanian \cite{joseph2020meta} proposed to incorporate the idea of meta learning in optimization. They pointed out that the distribution of the model parameters conditioned on a task can be seen as a meta distribution and proposed Meta-Consolidation for Continual Learning to learn this distribution with the help of the learned task-specific priors stored in the memory. 

Aside from the above methods in image classification, several works have also applied the memory based method to few-shot segmentation \cite{ganea2021incremental,shi2022incremental}. Among them, Ganea et al. \cite{ganea2021incremental} focused on few-shot instance segmentation. They proposed to replace the fixed feature extractor with an instance feature extractor, leading to more discriminative embeddings for each data instance. The embeddings of all instances from each previous class are averaged and then stored in the memory. Hence, given a new task, extensive retraining from scratch is not necessary. Shi et al. \cite{shi2022incremental} focused on few-shot semantic segmentation and also stored class-level embeddings in the memory. Further than that, during learning new-class embeddings, Shi et al. \cite{shi2022incremental} also proposed a strategy to seek help from existing embeddings, so that the learning process can less rely on the training data scale of the new classes. 

Besides the above two types of schemes to either retrain or optimize on the memory buffer, Prabhu et al. \cite{prabhu2020gdumb} proposed Greedy Sampler and Dumb Learner (GDumb), which consists of a greedy sampler and a dumb learner. GDumb greedily constructs the memory buffer from the data sequence with the objective of having a memory buffer with a balanced task distribution. At inference, GDumb simply trains the model from scratch on only the data in the memory buffer. The authors pointed out that even though GDumb is not particularly designed for continual learning, it outperforms several continual learning frameworks.

\subsection{Generative Replay Based Methods}

Generative replay based methods \cite{shin2017continual, kemker2017fearnet} have been proposed as an alternative to the memory based methods by replacing the memory buffer with a generative module that reproduces information related to the previous tasks. This can be achieved through a variety of schemes, which are introduced below.

\subsubsection{Generating Previous Data Instances}
Typically, most generative replay based methods \cite{shin2017continual, van2018generative} generate only previous data instances along with their corresponding labels. 

In image classification, Shin et al. \cite{shin2017continual} proposed a Deep Generative Replay (DGR) method, in which a GAN model is trained to generate previous data instances and pair them with their corresponding labels. Kemker and Kanan \cite{kemker2017fearnet} also proposed a similar method, which uses pseudo-rehearsal \cite{robins1995catastrophic} to enable the model to revisit recent memories. Van de Ven and Tolias \cite{van2018generative} pointed out that the generative replay based methods can perform well, especially when combined with knowledge distillation, but the computation cost is usually very heavy. Hence, they proposed an efficient generative replay based method by integrating the helper GAN model into the main model used for classification. Cong et al. \cite{cong2020gan} proposed a GAN memory to mitigate catastrophic forgetting by learning an adaptive GAN model. They adopted the modified variants of style-transfer techniques to transfer the base GAN model towards each task domain, leading to increased quality of the generated data instances. Besides the typical GAN models, various other generative models were also explored. 
Lesort et al. \cite{lesort2019generative} investigated the performance of different generative models including Variational AutoEncoder (VAE), conditional VAE, conditional GAN, Wasserstein GAN, and Wasserstein GAN with Gradient Penalty. Among these generative models, GAN was still found to have the best performance but it was also pointed out that all generative models including GAN struggle with more complex datasets. Rostami et al. \cite{rostami2019complementary} trained an encoder to map different tasks into a task-invariant Gaussian Mixture Model (GMM). After that, for each new task, the pseudo previous data instances will be generated from this GMM model through a decoder to be trained collaboratively with the current data instances. 

More recently, Wang et al. \cite{wang2021ordisco} focused on utilizing unlabelled data instances and proposed Continual Replay with Discriminator Consistency. A minimax adversarial game strategy \cite{goodfellow2014generative} was adapted where the output distribution of the generator was used to update the classifier and the output pseudo labels of the classifier were used to update the generator. 
Moreover, Gao and Liu \cite{gao2023ddgr} explored generating previous data instances via a diffusion model. Specifically, in their proposed method, guided by the classifier that has already been trained over previous tasks, the diffusion model can generate high-quality data instances for previous tasks. Later on, inspired by the adversarial attack techniques, Goswami et al. \cite{goswami2024resurrecting} further proposed a method called Adversarial Drift Compensation. In their method, adversarial data instances are generated to better approximate the semantic drift between the previous and current tasks.

Aside from image classification, Wu et al. \cite{wu2018memory} applied continual learning in class-conditional image generation. They proposed to generate previous data instances and train the model on both generated and new data instances. 
Besides, Cai and M{\"u}ller \cite{cai2023clnerf} focused on incorporating continual learning and the Neural Radiance Field (NeRF). They first pointed out that, since NeRF is naturally a good data generator, tackling continual learning in the context of NeRFs from the perspective of generative replay can be very suitable. Based on this, they proposed a generative replay based continual learning method for NeRFs and further incorporated their method with an instant neural graphics primitives \cite{muller2022instant} architecture.
More recently, Ding et al. \cite{ding2024coherent} for the first time explored continual learning in the context of temporal action segmentation. Specifically, in their work, on top of the existing generative replay based continual learning methods in the image context, they further introduced a temporal coherence variable for facilitating the generator to generate temporal-coherent videos.

\subsubsection{Generating Previous Data Instances along with Latent Representations}
A few recent works \cite{van2020brain, ye2020learning} have also proposed to generate both previous data instances and their latent representations to further consolidate previously learned knowledge. 

In image classification, Van de Ven et al. \cite{van2020brain} pointed out that generating previous data instances for problems with complex inputs (e.g., natural images) is challenging. Hence, they proposed to also generate latent representations of data instances via context-modulated feedback connection of the network. Ye and Bors \cite{ye2020learning} proposed a lifelong VAEGAN which also learns latent representations of previous data instances besides the typical generative replay. It learns both shared and task-specific latent representations to facilitate representation learning. 

Apart from the above-mentioned schemes, i.e., Generating previous data instances and generating previous data instances along with latent representations, Campo et al. \cite{campo2020continual} applied the generative replay based method on general frame predictive tasks without utilizing the generated data instances. More precisely, they proposed to use the low-dimension latent features in between the encoder and the decoder of a VAE, which can further be integrated with other sensory information using a Markov Jump Particle Filter \cite{baydoun2018learning}. 

\subsection{Parameter Isolation Based Methods}

Parameter isolation based methods \cite{rusu2016progressive, yoon2017lifelong, fernando2017pathnet, mallya2018packnet} generally assign different model parameters to different tasks to prevent later tasks from interfering with previously learned knowledge. This can be achieved through a variety of schemes, which are described below.

\subsubsection{Dynamic Network Architectures}

The majority of parameter isolation based methods \cite{rusu2016progressive, aljundi2017expert} assign different model parameters to different tasks by dynamically changing the network architecture. 

In image classification, in the early days, Rusu et al. \cite{rusu2016progressive} proposed a typical method called Progressive Network, which trains a new neural subnetwork for each new task. Through the training of each new subnetwork, feature transfer from subnetworks learned on the previous tasks is enabled by lateral connections. An analytical method based on the Fisher information matrix shows the effectiveness of this architecture in mitigating catastrophic forgetting. 
After that, Aljundi et al. \cite{aljundi2017expert} introduced a network of experts in which an expert gate was designed to only select the most relevant previous task to facilitate learning the new task. At test time, the expert gate structure is utilized to select the most approximate model for a given data instance from a certain task. 
Around the same time, Yoon et al. \cite{yoon2017lifelong} proposed a Dynamically Expandable Network that utilizes the knowledge learned from the previous tasks and expands the network structure when the previous knowledge is not enough to handle the new task. The addition, replication, and separation of neurons were designed to expand the network structure. 
Xu and Zhu \cite{xu2018reinforced} then further proposed a Reinforced Continual Learning method, which uses a recurrent neural network trained following an actor-critic strategy to determine the optimal number of nodes and filters that should be added to each layer for each new task. Later on, Li et al. \cite{li2019learn} proposed a Learn-to-Grow framework that searches for the optimal model architecture and trains the model parameters separately per task, while Hung et al. \cite{hung2019compacting} proposed a method called Compacting Picking Growing (CPG). For each new task, CPG freezes the parameters trained for all the previous tasks to prevent the model from forgetting any relevant information. Meanwhile, the new task is trained by generating a mask to select a subset of frozen parameters that can be reused for the new task and decide if additional parameters are necessary for the new task. After the above two steps, a compression process is conducted to remove the redundant weights generated by the training process of the new task.

As time passed, Lee et al. \cite{lee2020neural} pointed out that during human learning, task information is not necessary and hence proposed a Continual Neural Dirichlet Process Mixture model which can be trained both with a hard task boundary and in a task-free manner. It uses different expert subnetworks to handle different data instances. Creating a new expert subnetwork is decided by a Bayesian non-parametric framework. 
Hocquet et al. \cite{hocquet2020ova} proposed One-versus-All Invertible Neural Networks in which a specialized invertible subnetwork \cite{dinh2014nice} is trained for each new task. At the test time, the subnetwork with the highest confidence score on a test sample is used to identify the class of the sample. 
Kumar et al. \cite{kumar2021bayesian} proposed to build each hidden layer with the Indian Buffet Process \cite{griffiths2011indian} prior. They pointed out that the relation between tasks should be reflected in the connections they used in the network, and thus proposed updating the connections during the training process of every new task. 
Yan et al. \cite{yan2021dynamically} proposed a new method with a new data representation called super-feature. For each new task, a new task-specific feature extractor is trained while keeping the parameters of the feature extractors of the previous tasks frozen. The features from all feature extractors are then concatenated to form a super-feature which is passed to a classifier to assign a label. 
Zhang et al. \cite{zhang2021few} proposed a Continually Evolved Classifier (CEC) focusing particularly on the few-shot scenario of continual learning. They first proposed to separate representation and classification learning, where representation learning is frozen to avoid forgetting, and the classification learning part is replaced by a CEC which uses a graph model to adapt the classifier towards different tasks based on their diverse context information. 

Moreover, particularly from the efficiency perspective, Veniat et al. \cite{veniat2020efficient} built an efficient continual learning model, whose architecture consists of a group of modules representing atomic skills that can be combined in different ways to perform different tasks. When a new task comes, the model potentially reuses some of the existing modules and creates a few new modules. The decision of which modules to reuse and which new modules to create is made by employing a data-driven prior. As an extension of \cite{veniat2020efficient}, Valkov et al. \cite{valkov2024a} pointed out that, in the modular structure utilized in \cite{veniat2020efficient}, the number of possible module compositions can grow quite quickly along with the growth of the number of tasks. Thus, given a new task, how to properly compute a module composition's fitness over the task then becomes a key issue. Considering the above, Valkov et al. \cite{valkov2024a} proposed PICLE, which tackles this issue via a probabilistic search \cite{shahriari2015taking}. 

Several other approaches \cite{yoon2019scalable,kanakis2020reparameterizing,ebrahimi2020adversarial} have focused on separating the shared and task-specific features to effectively prevent new tasks from interfering with previously learned knowledge. Yoon et al. \cite{yoon2019scalable} proposed a method called Additive Parameter Decomposition (APD). APD first uses a hierarchical knowledge consolidation method to construct a tree-like structure for the model parameters on the previous tasks where the root represents the parameters that are shared among all the tasks and the leaves represent the parameters that are specific to one task. The middle nodes of the tree represent the parameters that are shared by a subset of the tasks. Given a new task, APD tries to encode the task with shared parameters first and then expands the tree with task-specific parameters. Kanakis et al. \cite{kanakis2020reparameterizing} proposed Reparameterized Convolutions for Multi-task learning which separates the model parameters into filter-bank part and modulator part. The filter-bank part of each layer is pre-trained and shared among tasks encoding common knowledge, while the modulator part is task-specific and is separately fine-tuned for each new task.
Ebrahimi et al. \cite{ebrahimi2020adversarial} trained a common shared module to generate task-invariant features. A task-specific module is expanded for each new task to generate task-specific features orthogonal to the task-invariant features. Singh et al. \cite{singh2020calibrating} separated the shared and the task-specific components by calibrating the activation maps of each layer with spatial and channel-wise calibration modules which can adapt the model to different tasks. Singh et al. \cite{singh2021rectification} further extended \cite{singh2020calibrating} to be used for both zero-shot and non-zero-shot continual learning. 
Verma et al. \cite{verma2021efficient} proposed an Efficient Feature Transformation to separate the shared and the task-specific features using Efficient Convolution Operations \cite{howard2017mobilenets}. 

More recently, with the emergence of prompt tuning \cite{smith2023coda,wang2022dualprompt,wang2022learning,gao2023unified}, the convolution operation is also utilized by Roy et al. \cite{roy2024convolutional} over the task-shared features to generate task-specific prompts for each new task. Moreover, Gao et al. \cite{gao2024consistent} also focused on prompt-tuning-based continual learning yet from another perspective. Specifically, they pointed out that in existing prompt-tuning-based methods, an inconsistency can exist between training and testing. Correspondingly, they proposed a Consistent Prompting method to tackle this problem. Around the same time, Yu et al. \cite{yu2024boosting} explored over prompt-tuning-based continual learning as well. In their proposed method, a Mixture-of-Experts adapter structure is first incorporated, and a Distribution Discriminative Auto-Selector is further designed for routing the input data instance into either the adapter or the original vision-and-language model (i.e., CLIP). Moreover, the prompt-tuning-based continual learning method is also explored in \cite{liang2024inflora}, in which a new way of injecting task-specific prompts to the model named InfLoRA was proposed. Specifically, Liang and Li \cite{liang2024inflora} analyzed that, by their way of injecting prompts, for each task, optimizing its task-specific prompt is equivalent to optimizing the model in a task-specific subspace. Based on this analysis, Liang and Li \cite{liang2024inflora} concluded that, InfLoRA can effectively mitigate the interference across different tasks. Besides the above, recently, Kim et al. \cite{kim2024one} also proposed another prompt-tuning-based continual learning named AdaPromptCL. In their method, based on the different degrees of semantic shifts between different tasks, a new task is learned either via updating an existing prompt or learning a new one.

Aside from image classification, a few recent works \cite{zhai2020piggyback, zhai2021hyper} have focused on the conditional image generation problem. Zhai et al. \cite{zhai2020piggyback} pointed out that not all parameters of a single Lifelong GAN \cite{zhai2019lifelong} can be adapted to different tasks. They thus proposed Piggyback GAN which has a filter bank containing the filters from layers of the model trained on the previous tasks. Given a new task, Piggyback GAN is learned to reuse a subset of filters in the filter bank and update the filter bank by adding task-specific filters to ensure high-quality generation of the new task. It was pointed out by Zhai et al. \cite{zhai2021hyper} that this method is not memory efficient. Hence, Hyper-LifelongGAN was proposed to apply a hypernetwork \cite{ha2016hypernetworks} to the lifelong GAN framework. Instead of learning a deterministic combination of filters for each task, the hypernetwork learns a dynamic filter as task-specific coefficients to be multiplied with the task-independent base weight matrix.

Further to this, some works have focused on other computer vision problems, such as image generation \cite{seo2023lfs}, action recognition \cite{li2021elsenet}, multi-modal reasoning \cite{xu2023experts}, label-to-image translation \cite{chen2022few}, and multi-task learning \cite{hung2019increasingly}. 
Among them, Seo et al. \cite{seo2023lfs} focused on the task of few-shot image generation and proposed a method named Lifelong Few-Shot GAN. In their proposed method, for each task, an efficient task-specific modulator is learned to modulate the weights of the model correspondingly. 
Li et al. \cite{li2021elsenet} proposed an Elastic Semantic Network for skeleton-based human action recognition, which comprises a base model followed by a stack of multiple elastic units. Each elastic unit consists of multiple learning blocks and a switch block. For every new action as a task, only the most relevant block in each layer is selected by the switch block to facilitate the training process while all the other blocks are frozen to preserve previously learned knowledge. 
Xu and Liu \cite{xu2023experts} explored continual learning in the multi-modal learning area and proposed an Experts collaboration network (Expo). Expo first incorporates separate groups of experts for each modality so that inputs from different modalities can all be effectively processed. Moreover, for new skills to be effectively acquired, Expo also designs its expert groups to be expandable.
Chen et al. \cite{chen2022few} explored continual learning in the context of label-to-image translation. In their proposed method, a set of class-adaptive scale, shift, and bias parameters are first learned for every new class. These parameters are then used for base convolution and normalization for this class.
Hung et al. \cite{hung2019increasingly} proposed a Packing-And-Expanding (PAE) method that sequentially learns face recognition, facial expression understanding, gender identification, and other relevant tasks in a single model. PAE improves Packnet \cite{mallya2018packnet} in two aspects. Firstly, PAE adapts iterative pruning procedure \cite{zhu2017prune} to construct a more compact model compared to Packnet. Secondly, PAE allows the model architecture to be expanded when the remaining trainable parameters are not enough to effectively learn the new task.
More recently, Kim et al. \cite{kim2024eclipse} further explore parameter isolation based continual learning in continual panoptic segmentation. Specifically, similar to \cite{gao2024consistent,roy2024convolutional}, Kim et al. \cite{kim2024eclipse} focused on prompt-tuning-based continual learning, while they further designed a logit manipulation strategy, particularly for the error propagation and semantic drift problems in continual panoptic segmentation.

\subsubsection{Fixed Network Architectures}

\begin{figure}
  \includegraphics[width=\linewidth]{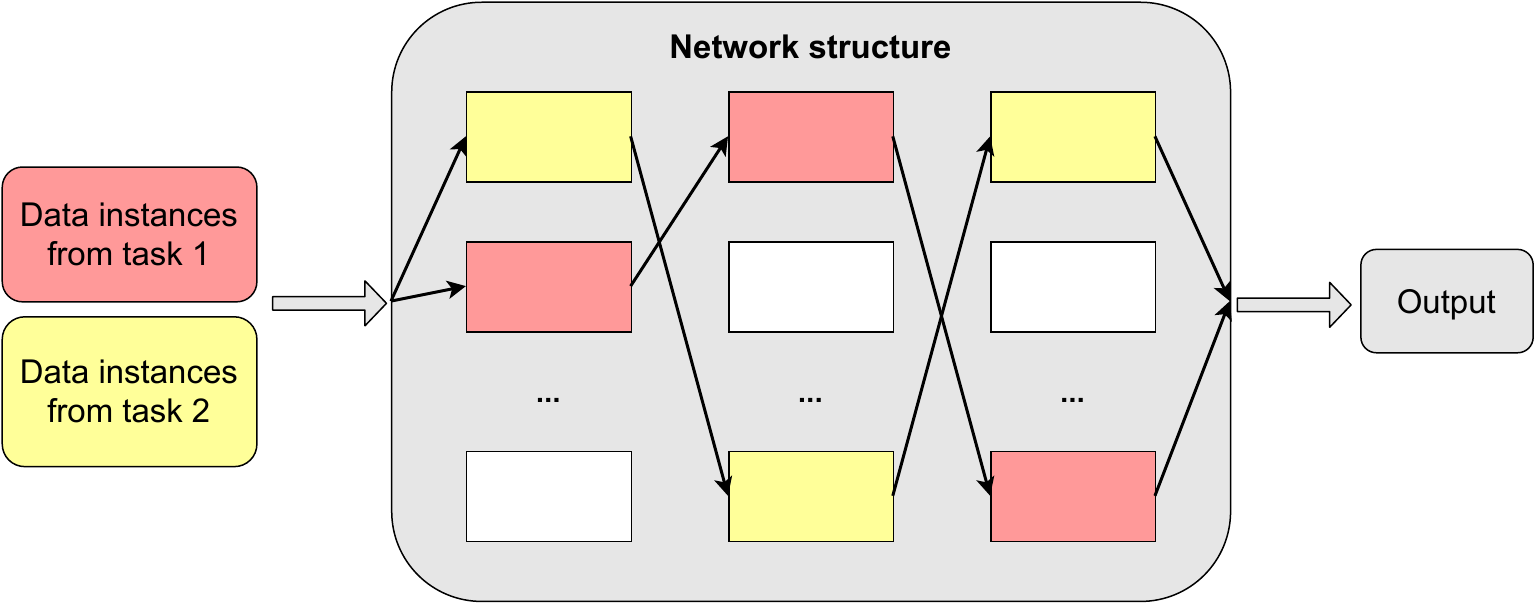}
  \caption{Illustration of PathNet \cite{fernando2017pathnet}, which freezes the model parameters along all paths selected by the previous tasks and reinitializes and retrains the remaining model parameters following the same process to select the best path.}
  \label{fig:pi}
\end{figure}

Other than using parameter isolation based methods to dynamically change the network architectures, several other works \cite{fernando2017pathnet, serra2018overcoming} have designed fixed network architectures that can still assign different parameters for handling different tasks. 

In image classification, Fernando et al. \cite{fernando2017pathnet} proposed PathNet. In the training of the first task, several random paths through the network are selected. After that, a tournament selection genetic algorithm is utilized to select the best path to be trained for the task. Then for each following task, as shown in Figure \ref{fig:pi}, model parameters along all paths selected by the previous tasks are frozen and the remaining parameters are re-initialized and trained again following the above process. Serra et al. \cite{serra2018overcoming} proposed Hard Attention to the Task (HAT), which calculates an almost binary attention vector for each task during training. For each following task, all the attention vectors calculated before are used as the masks to freeze the network parameters which are crucial for the previous tasks. A similar idea was also proposed by Mallya et al. \cite{mallya2018piggyback}. Rajasegaran et al. \cite{rajasegaran2019random} pointed out that the PathNet \cite{fernando2017pathnet} requires heavy computational costs. Hence, they proposed to reuse paths of the previous tasks to achieve a much lower computational overhead.
From a similar perspective, Rype{\'s}{\'c} et al. \cite{rype2024divide} also worked on reducing the computational overhead. Specifically, for each new task, rather than optimizing all the experts in a computationally heavy manner, with the help of the multivariate Gaussian distributions, they enabled only a single expert to be selected and correspondingly optimized.
Abati et al. \cite{abati2020conditional} proposed a gating module for each convolution layer to select a limited set of filters and protect the filters of the previous tasks from being unnecessarily updated. A sparsity objective was also used to make the model more compact. 
Unlike \cite{fernando2017pathnet,serra2018overcoming,abati2020conditional}, Shi et al. \cite{shi2021continual} proposed a Bit-Level Information Preserving method, which updates the model parameters at the bit level by estimating information gain on each parameter. A certain number of bits are frozen to preserve information gain provided by the previous tasks. 

Besides the above-mentioned methods, freezing a subset of the model parameters with the help of meta learning has also received attention recently. Beaulieu et al. \cite{beaulieu2020learning} proposed a Neuromodulated Meta-Learning Algorithm that contains a neuromodulatory neural network and a prediction learning network. The neuromodulatory neural network is meta-trained to activate only a subset of parameters of the prediction learning network for each task to mitigate catastrophic forgetting.
Hurtado et al. \cite{hurtado2021optimizing} proposed a method called Meta Reusable Knowledge (MARK), which has a single common knowledge base for all the learned tasks. For each new task, MARK uses meta learning to update the common knowledge base and uses a trainable mask to extract the relevant parameters from the knowledge base for the task.   

Unlike the above-mentioned methods that focused on how to freeze a subset of model parameters per task, Mallya and Lazebnik \cite{mallya2018packnet} proposed PackNet, which first trains the whole neural network on each new task and then uses a weight-based pruning technique to free up unnecessary parameters. Adel et al. \cite{adel2019continual} proposed Continue Learning with Adaptive Weights (CLAW), which adapts the model parameters in different scales in a data-driven way. More precisely, CLAW is a variational inference framework that trains three new groups of parameters for each neuron, including one binary parameter to represent whether or not to adapt the neuron, and the two other parameters to represent the magnitude of adaptation.

Besides, Hu et al. \cite{hu2018overcoming} proposed a Parameter Generation and Model Adaptation method, in which the model parameters are dynamically changed at inference time instead of training time. The model includes both a set of shared parameters and a parameter generator. At inference time, the parameter generator with the test data instance as input, outputs another set of parameters. After that, the shared set and the output set of parameters are used together to classify the test data instance. 

Aside from image classification, some works have focused on other computer vision problems, such as interactive image segmentation \cite{zheng2021continual} and image captioning \cite{del2020ratt}. 
Zheng et al. \cite{zheng2021continual} proposed to activate only a subset of convolutional kernels by employing a Bayesian non-parametric Indian Buffet Process \cite{griffiths2011indian}, which results in extracting the most discriminative features for each task. The kernels that are frequently activated in the previous tasks are encouraged to be re-activated for the new tasks to facilitate knowledge transfer.
Chiaro et al. \cite{del2020ratt} applied the parameter isolation based method on image captioning by extending the idea of HAT \cite{serra2018overcoming} to recurrent neural networks. They pointed out that image captioning allows the same words to describe images from different tasks. Hence, compared to HAT, they allowed the vocabulary masks for different tasks to have shared elements.

\subsection{Combination of Multiple Categories of Methods}
To enhance the performance of continual learning, several other works proposed to combine two or more categories of the aforementioned techniques.

\subsubsection{Combination of Regularization and Memory Based Methods}

Several approaches \cite{nguyen2017variational, aljundi2019task} take advantage of both the regularization and memory based methods to enhance continual learning performance in image classification. Nguyen et al. \cite{nguyen2017variational} proposed Variational Continual Learning (VCL) which combines variational inference and coreset data summarization \cite{bachem2015coresets} together. VCL recursively calculates the posterior information of the existing model from a subset of important previous data instances and merges it with the likelihood information calculated from the new task. Similarly, Kurle et al. \cite{kurle2019continual} used both the posterior information and a subset of previous data instances. However, they introduced a new update method and further adapted the Bayesian neural network to non-stationary data. Titsias et al. \cite{titsias2019functional} pointed out that previous methods such as EWC \cite{kirkpatrick2017overcoming} and VCL \cite{nguyen2017variational} suffer from the problem of brittleness as a subset of model parameters are preserved towards the previous tasks while the other parameters are further trained towards the new task. They thus proposed a functional regularisation approach that marginalizes out the task-specific weights to resolve the problem of brittleness. They also proposed to store a subset of inducing points for each task. When a new task comes, these stored inducing points are used to conduct functional regularisation to mitigate catastrophic forgetting. Chen et al. \cite{chen2021overcoming} pointed out that the assumption made by VCL \cite{nguyen2017variational}, i.e., the shared and the task-specific parameters are independent of each other, is not true in general. Hence, they proposed to use an energy-based model \cite{lecun2006tutorial} together with Langevin dynamic sampling \cite{bussi2007accurate} as a regularizer to choose the shared parameters that satisfy the independent assumption of VCL.

Besides VCL \cite{kirkpatrick2017overcoming} and its extensions, various other image classification methods \cite{riemer2018learning, aljundi2019task} have also been proposed. 
Riemer et al. \cite{riemer2018learning} integrated a modified version of the Reptile algorithm \cite{nichol2018reptile} into an experience replay method to effectively mitigate catastrophic forgetting and preserve the ability of future learning. 
Aljundi et al. \cite{aljundi2019task} proposed a method that allows classes to be repeated between different tasks. They utilized the structure of the memory-aware synapse \cite{aljundi2018memory} as a weight regularizer and stored hard data instances with the highest loss to help identification of important parameters of the previous tasks.  
Similar to \cite{titsias2019functional}, Pan et al. \cite{pan2020continual} also employed functional regularisation to only regularize the model outputs. 
But, unlike \cite{titsias2019functional} which requires solving a discrete optimization problem to select inducing points, they proposed to store data instances that are close to the decision boundary.
Tang and Matteson \cite{tang2020graph} generated random graphs from the stored data instances and introduced a new regularization term penalizing the forgetting of the edges of the generated graphs. Mirzadeh et al. \cite{mirzadeh2020linear} pointed out that there exists a linear connection between continual learning and multi-task learning. Based on this, they proposed Mode Connectivity Stochastic Gradient Descent to treat continual learning as multi-task learning. A memory based method is employed to approximate the loss of the previous tasks. 

More recently, Zhang et al. \cite{zhang2024continual} explored the budget continual learning setting \cite{prabhu2023computationally} and further extended it to a semi-supervised manner. Specifically, in their proposed method, for each new task, besides using the labeled data instances from this task, they also utilize both the stored labeled data instances from the previous tasks for task balancing, and the unlabelled data instances from the current task for regularization. Besides, a budget allocation stage is also involved to further balance the current and previous tasks. Around the same time, Magistri et al. \cite{magistri2024elastic} also explored the combination of regularization and memory based methods yet from another perspective. Specifically, they performed their regularization with the help of a positive semi-definite Empirical Feature Matrix, while at the same time proposing an Asymmetric Prototype Replay loss for better balancing the stored prototypes from the previous tasks and the data instances from the new task.

\subsubsection{Combination of Knowledge Distillation and Memory Based Methods}

A few other approaches \cite{castro2018end, hou2018lifelong} have combined the knowledge distillation and memory based methods. In image classification, Castro et al. \cite{castro2018end} proposed a cross-distilled loss consisting of a knowledge distillation loss to preserve the knowledge acquired from the previous tasks, and a cross-entropy loss to learn the new task. They also introduced a representative memory unit that performs a herding-based selection and removal operation to update the memory buffer. Hou et al. \cite{hou2018lifelong} pointed out that with the help of knowledge distillation, storing only a small number of previous data instances can lead to a large improvement in mitigating catastrophic forgetting. They thus proposed a method named Distillation and Retrospection to obtain a better balance between the preservation of previous tasks and adaptation to new tasks. 

More recently, Yan et al. \cite{yan2024orchestrate} pointed out that, particularly in online continual learning, many existing methods can suffer from the overfitting-underfitting dilemma. That is, they can underfit to the data instances from the new task, while at the same time overfit to those stored data instances from the previous tasks. To tackle this problem, in the work of Yan et al. \cite{yan2024orchestrate}, both a multi-level supervision scheme and a reverse self-distillation technique are proposed. Wang et al. \cite{wang2024improving} also focused on online continual learning yet from the plasticity perspective. Specifically, according to Wang et al. \cite{wang2024improving}, existing memory-based online continual learning methods, while can mitigate catastrophic forgetting to some extent, can still suffer from lacking model plasticity (i.e., lacking the ability to acquire new knowledge). To tackle this issue, on top of the memory-based methods, they further proposed to use collaborative learning \cite{anil2018large} and distillation chain to improve the plasticity of the continual learning models.

In video classification, Zhao et al. \cite{zhao2021video} adapted the knowledge distillation based method by separately distilling the spatial and temporal knowledge. They also introduced a dual granularity exemplar selection method to store only key frames of representative video instances from the previous tasks. 

In fake media detection, Kim et al. \cite{kim2021cored} proposed to store the feature representations of both real and fake data instances to facilitate the knowledge distillation process. 

In object detection, Liu et al. \cite{liu2023augmented} first identified foreground shift as a crucial problem for memory based methods in continual object detection. They then proposed a method named Augmented Box Replay to handle this problem, and a loss function named attentive Region-of-Interest distillation loss to further reduce catastrophic forgetting. 

In embodied agent learning, inspired by previous works \cite{buzzega2020dark,boschini2022class}, Kim et al. \cite{kim2024online} also took the logit distillation idea to enable the embodied agents to be learned continually. Yet different from previous works, to handle the outdated logit issue while at the same time releasing the agent's requirement on the task boundary information, Kim et al. \cite{kim2024online} further proposed a Confidence-Aware Moving Average method.

Moreover, recently, Ye et al. \cite{ye2024continual} explored continual learning w.r.t. multi-modal medical data as well. Specifically, in their proposed method, they first utilize a k-means sampling strategy to sample and store data instances from the previous tasks. After that, during the learning process of a new task, a feature distillation strategy and an intra-modal mixup strategy are then utilized on those stored data.

\subsubsection{Combination of Knowledge Distillation and Generative Replay Based Methods}

The combination of knowledge distillation and generative replay based methods has also been explored \cite{wu2018incremental,zhang2023target,huang2021half}.
In image classification, Wu et al. \cite{wu2018incremental} proposed a new loss function that consists of a cross-entropy loss and a knowledge distillation loss. A GAN is employed to generate data instances from the previous tasks, which are then combined with real new data instances to train the model. A scalar, which represents the bias on data instances from the new task, is also used to remove the bias caused by data imbalance. 
Besides Wu et al. \cite{wu2018incremental}, Zhang et al. \cite{zhang2023target} also explore continual learning in image classification but they particularly focus on federated continual learning. Specifically, in their method, a generator is first trained to generate data instances that can simulate the global distribution of previous tasks. After that, a knowledge distillation process is further involved, in which the student model is utilized to further enhance the capacity of the generator. 
Similar to \cite{zhang2023target}, Abudukelimu et al. \cite{wuerkaixi2024accurate} also focused on federated continual learning and included a knowledge distillation process in their method. Yet different from \cite{zhang2023target} which performs data instance generation, Abudukelimu et al. \cite{wuerkaixi2024accurate} proposed to learn a normalizing flow model \cite{durkan2019neural} and performed feature generation. Moreover, a correlation estimation process is further involved in their method to forget those biased features from the previous tasks.

In image semantic segmentation, Huang et al. \cite{huang2021half} proposed to generate fake images of the previous classes using a Scale-Aware Aggregation module and combine them with the images from the new task to facilitate the knowledge distillation process.
Moreover, recently, Kim et al. \cite{kim2024sddgr} focused on the application of continual learning in object detection. Specifically, in their method, the pre-trained text-to-image diffusion model is utilized for data instance synthesis, while an L2 knowledge distillation scheme is further leveraged to ease the knowledge transfer process from the synthesized data instances toward the continual learning model.

\subsubsection{Combination of Memory and Generative Replay Based Methods}

A few other works \cite{he2018exemplar, xiang2019incremental} have proposed to combine the memory and generative replay based techniques.
In image classification, He et al. \cite{he2018exemplar} proposed an Exemplar Supported Generative Reproduction (ESGR) method, which leverages both generated data instances and stored real data instances to better mitigate catastrophic forgetting. Specifically, given a new task, ESGR first generates data instances of previous tasks by employing task-specific GANs. Then, the combination of the stored previous data instances, the generated instances, and new data instances is used to train the classification model. 
Unlike ESGR \cite{he2018exemplar} which requires real previous data instances to be stored, Xiang et al. \cite{xiang2019incremental} propose storing statistical information of previous data instances, such as mean and covariance, which are fed to a conditional GAN to generate pseudo data instances. 
Ayub and Wagner \cite{ayub2021eec} proposed a memory-efficient method named Encoding Episodes as Concepts (EEC). When there still exists enough memory space, EEC stores compressed embeddings of data instances in the memory buffer and retrieves these embeddings while learning new tasks. When memory space is full, EEC combines similarly stored embeddings into centroids and covariance matrices to reduce the amount of memory required. It then generates pseudo data instances from these matrices while learning new tasks.

More recently, as an extension of \cite{yang2023neural} yet in the online continual learning area, Seo et al. \cite{seo2024learning} proposed a method named EARL. Specifically, in EARL, during the learning process of a new task, the stored data instances, as well as the preparatory data instances generated leveraging hard transform from those stored data instances, are both utilized. Particularly, according to the authors, the utilization of the preparatory data instances can avoid biased predictions towards the previous tasks. In image semantic segmentation, Wu et al. \cite{wu2019ace} proposed to store feature statistics of previous images to help train the image generator. The trained image generator is then used to align the style of previous and new coming images by matching their first and second-order feature statistics.

\subsubsection{Combination of Generative Replay and Parameter Isolation Based Methods}

A few other methods \cite{ostapenko2019learning, rao2019continual} explore the combination of generative replay and parameter isolation based methods for image classification. Ostapenko et al. \cite{ostapenko2019learning} combined the HAT method \cite{serra2018overcoming} with the generative replay based method using a modified auxiliary classifier GAN \cite{odena2017conditional} architecture. Rao et al. \cite{rao2019continual} proposed a new problem setup called unsupervised continual learning, in which information including task labels, task boundaries, and class labels are unknown. To solve this problem, they proposed a Continual Unsupervised Representation Learning (CURL) framework, which can expand itself dynamically to learn new concepts. Besides, CURL also involves the mixture generative replay as an extension of DGR \cite{shin2017continual} to conduct unsupervised learning without forgetting. More recently, Zhou et al. \cite{zhou2024expandable} further proposed a pre-trained-model-based continual learning method called EASE. Specifically, in their method, for each new task, a task-specific adapter is first trained to formulate a task-specific feature subspace. After that, with the prototypes of the previous tasks synthesized and the prototype of the current task extracted, a subspace ensemble process is then performed to build the full classifier across all tasks.

\subsubsection{Other Combinations}
Apart from the above-mentioned combination approaches, there are several other methods \cite{smith2021memory, yang2019adaptive, buzzega2020dark} which proposed other combinations to perform image classification or other computer vision tasks. 
In image classification, Smith et al. \cite{smith2021memory} combined regularization and knowledge distillation based methods. They proposed DistillMatch to conduct both local soft distillation and global hard distillation using unlabelled data. Besides, a consistency regularization was also proposed which uses the consistency loss \cite{sohn2020fixmatch} to increase the number of robust decisions made. 
Yang et al. \cite{yang2019adaptive} combined regularization and parameter isolation based methods. They proposed an Incremental Adaptive Deep Model (IADM), which has an attention module for the hidden layers of the model, enabling training of the model with adaptive depth. IADM also incorporates adaptive Fisher regularization to calculate the distribution of both the previous and new data instances to mitigate catastrophic forgetting of the previous tasks. 
Buzzega et al. \cite{buzzega2020dark} combined regularization, knowledge distillation, and memory based methods and proposed the Dark Experience Replay (DER). Compared to ER \cite{chaudhry2019tiny}, DER further distills previously learned knowledge with the help of Dark Knowledge \cite{hinton2014dark} and regularizes the model based on the network's logits so that consistent performance can be expected through the sequential stream of data without a clear task boundary. 

More recently, Zhou and Hua \cite{zhou2024defense} further explored the setting of continual adversarial defense under a sequence of attacks, and combined regularization, knowledge distillation, and generative replay based methods in their work. Specifically, in their proposed method, a set of data instances is first generated through both isotropic and anisotropic pseudo-replay mechanisms. After that, these generated data instances are utilized to optimize the model through a self-distillation loss, while a regularization term is also further incorporated into the proposed framework to better tackle the stability-plasticity dilemma. 

\subsection{Other Methods}

Aside from the above-mentioned categories of methods, several other techniques \cite{caccia2020online, yu2020self} have also been proposed for continual learning. 

In image classification, several meta learning based methods have been proposed. Caccia et al. \cite{caccia2020online} introduced a new scenario of continual learning, where the model is required to solve new tasks and remember the previous tasks quickly. Thus, continual-MAML, i.e. an extension of Model-Agnostic Meta-Learning (MAML) \cite{finn2017model}, was proposed to first initialize the model parameters with meta learning and then add new knowledge into the learned initialization only when the new task has a significant distribution shift from the previously learned tasks. Javed and White \cite{javed2019meta} proposed a meta-objective to learn representations that are naturally highly sparse and thus effectively mitigate catastrophic forgetting in continual learning. Jerfel et al. \cite{jerfel2018reconciling} employed a meta-learner to control the amount of knowledge transfer between tasks and automatically adapt to a new task when a task distribution shift is detected. 
Rajasegaran et al. \cite{rajasegaran2020itaml} proposed an incremental Task-Agnostic Meta-learning (iTAML) method, which adapts meta learning to separate the task-agnostic feature extractor from the task-specific classifier. In this way, iTAML first learns a task-agnostic model to predict the task, and then adapts to the task in the second step. This two-phase learning enables iTAML to mitigate the data imbalance problem by updating the task-specific parameters separately for each task.

There are also several generic frameworks that can be integrated with methods in the above-mentioned continual learning categories to further improve their performance. Yu et al. \cite{yu2020semantic} pointed out that embedding networks compared to the typical classification networks suffer much less from catastrophic forgetting. Hence, they adapted LwF \cite{li2017learning}, EWC \cite{kirkpatrick2017overcoming}, and MAS \cite{aljundi2018memory} towards embedding networks. Liu et al. \cite{liu2020more} pointed out that various classifiers with different decision boundaries can help to mitigate catastrophic forgetting. Hence, they proposed a generic framework called MUlti-Classifier (MUC) to train additional classifiers for each task, particularly on out-of-distribution data instances. They showed that MUC can further mitigate catastrophic forgetting when combined with other methods such as MAS \cite{aljundi2018memory} and LwF \cite{li2017learning}. Mendez and Eaton \cite{mendez2020lifelong} incorporated compositional learning into continual learning and proposed a generic framework that is agnostic to any continual learning algorithm. The framework consists of several components including linear models, soft layer ordering \cite{meyerson2017beyond}, and soft gating as a modified version of soft layer ordering, which can be combined together in different orders to construct different models for different tasks. Given a new task, the model first uses the previously learned components to solve the task, and then updates the components and creates new components when necessary. 

More recently, Zou et al. \cite{zou2024compositional} also focused on the incorporation of compositional learning into continual learning. Yet, they particularly focused on few-shot continual learning, and correspondingly proposed a Primitive Reuse design. With this design, given a new continual-learning task, rather than creating new components, the task can be learned by just reusing those components from the base training stage. Around the same time, inspired by neuroscience principles, Wang et al. \cite{wang2024a} proposed another generic method named refresh learning. Specifically, to prevent a model from over-memorizing the information that is not really important and helpful to the previous tasks, the method proposes to process the model through an unlearn-relearn mechanism to first unlearn such unimportant information during the learning process of every new task.

Several other continual learning methods have also been proposed to perform image classification from different perspectives. Stojanov et al. \cite{stojanov2019incremental} proposed a synthetic incremental object learning environment called Continual Recognition Inspired by Babies, which enables generating various amounts of repetition. They then showed that continual learning with repetition is important in mitigating catastrophic forgetting. Wu et al. \cite{wu2021incremental} proposed ReduNet which adapts the newly proposed ''white box" deep neural network derived from the rate reduction principle \cite{chan2020deep} to mitigate catastrophic forgetting in continual image classification. Knowledge from the previous tasks is explicitly preserved in the second-order statistics of ReduNet. 
Chen et al. \cite{chen2020long} adapted the lottery ticket hypothesis \cite{frankle2018lottery} to continual learning. They proposed a bottom-up pruning method to build a sparse subnetwork for each task, which is lightweight but can still achieve comparable or even better performance than the original dense model. 
Zhu et al. \cite{zhu2021self} mapped the representations of the previous tasks and the new task into the same embedding space and further used their distance to guide the model to preserve previous knowledge. A randomly episodic training and self-motivated prototype refinement were also proposed to extend the representation ability of the feature space given only a few data instances per task.
Liu et al. \cite{liu2021adaptive} proposed Adaptive Aggregation Networks which add a stable block and a plastic block to each residual level and aggregate their outputs to balance the ability to learn new tasks while remembering previous tasks.
Abdelsalam et al. \cite{abdelsalam2021iirc} proposed a new problem setup called Incremental Implicitly-Refined Classification, in which each data instance has a coarse label and a fine label. This problem setup is closer to real-world scenarios as humans interact with the same family of entities multiple times and discover more granular information about them over time, while trying not to forget previous knowledge. Several continual learning approaches have been evaluated on this setup.

Aside from image classification, Yu et al. \cite{yu2020self} introduced a continual learning method for image semantic segmentation. Before the training process of each new task, the model containing previous knowledge is stored. After training a new model on data instances from the new task, a set of unlabelled data instances are fed into both the previous and the new models to generate their pseudo labels, which are then fused by a Conflict Reduction Module to get their most accurate pseudo labels. Finally, a joint model is retrained with the unlabelled data instances and their corresponding pseudo labels to preserve previously learned knowledge. 
Besides, Wang et al. \cite{wang2022lifelong} proposed a new continual learning framework for graph learning named Feature Graph Networks (FGN). Specifically, FGN designs a new graph topology through which the continual growth of a graph can be regarded as the continual increase of training samples, and typical convolutional-neural-network-based continual learning techniques can thus be used for graph learning. 

More recently, Liu et al. \cite{liu2024locality} further explored continual learning under the model-based reinforcement learning setting, in an analytical manner similar to previous works \cite{zhuang2022acil,zhuang2023gkeal}. Yet different from \cite{zhuang2022acil,zhuang2023gkeal}, Liu et al. \cite{liu2024locality} proposed to incorporate their framework with a novel Losse feature extractor, which ensures feature sparsity while at the same time releasing the framework from a pre-training step. 
Also at a recent time, Garg et al. \cite{garg2024ticclip} formulated the TIC-DataComp benchmark, which according to the authors, serves as a pioneering exploration of continual learning in the context of large-scale vision-language models like CLIP \cite{radford2021learning}. 
Moreover, Liu et al. \cite{liu2024vida} focused on the continual test time adaptation task \cite{wang2022continual} and proposed a method named ViDA. Specifically, in ViDA, the domain-specific knowledge and the domain-shared knowledge are first explicitly managed through a high-rank branch and a low-rank branch in the model. After that, for each data instance, knowledge from different branches is then further merged adaptively via a Homeostatic Knowledge Allotment strategy.  
Besides, Carreira et al. \cite{carreira2024learning} explored the setting of continual learning in a single video stream, in which video frames as data instances can be highly correlated. In such a scenario, Carreira et al. \cite{carreira2024learning} further made an interesting observation that the usage of momentum during optimization seems to be unhelpful.
In addition, Pan et al. \cite{pan2024adaptive} further designed an Adaptive Visual-Inertial Odometry framework. With this framework, they enabled online continual learning to be incorporated into Visual-Inertial Odometry, via designing a factor graph optimization scheme and enabling it to provide feedback signals to the networks.

\section{Performance Comparison}

In the previous section, we reviewed recent continual learning methods across different computer vision tasks. Here, we will further present the performance of different continual learning methods across a variety of different computer vision tasks. Specifically, below, over each discussed task, we will first introduce the datasets and experimental settings that continual learning methods in this task commonly use and follow. The performance comparison across different continual learning methods in this task is then shown. Note that for the image classification task in which lots of different continual learning methods have been proposed, we focus on the benchmarks of online continual learning and few-shot class incremental learning, which serve as benchmarks that especially received lots of research attention recently.

\begin{table*}
\centering
\caption{Performance comparison among different methods under the online continual learning setting.}
\resizebox{\textwidth}{!}{%
\begin{tabular}{r|cccccc|cccccc}
\toprule
Dataset &
  \multicolumn{6}{c|}{CIFAR-100} &
  \multicolumn{6}{c}{Tiny-ImageNet} \\ \midrule
Memory Size &
  \multicolumn{2}{c|}{1,000} &
  \multicolumn{2}{c|}{2,000} &
  \multicolumn{2}{c|}{5,000} &
  \multicolumn{2}{c|}{2,000} &
  \multicolumn{2}{c|}{4,000} &
  \multicolumn{2}{c}{10,000} \\ \midrule 
Evaluation Metric &
  $A_q$(\%) $\uparrow$ &
  \multicolumn{1}{c|}{{ $F_q$(\%) $\downarrow$}} &
  $A_q$(\%) $\uparrow$ &
  \multicolumn{1}{c|}{{ $F_q$(\%) $\downarrow$}} &
  $A_q$(\%) $\uparrow$ &
  { $F_q$(\%) $\downarrow$} &
  $A_q$(\%) $\uparrow$ &
  \multicolumn{1}{c|}{{ $F_q$(\%) $\downarrow$}} &
  $A_q$(\%) $\uparrow$ &
  \multicolumn{1}{c|}{{ $F_q$(\%) $\downarrow$}} &
  $A_q$(\%) $\uparrow$ &
  { $F_q$(\%) $\downarrow$} \\ \midrule \midrule
Chaudhry et al. \cite{chaudhry2018efficient} &
  5.8$\pm$0.2 &
  \multicolumn{1}{c|}{{ 77.6$\pm$2.0}} &
  5.9$\pm$0.3 &
  \multicolumn{1}{c|}{{ 76.9$\pm$1.5}} &
  6.1$\pm$0.4 &
  { 78.3$\pm$1.2} &
  0.9$\pm$0.1 &
  \multicolumn{1}{c|}{{ 73.9$\pm$0.2}} &
  2.0$\pm$0.5 &
  \multicolumn{1}{c|}{{ 77.9$\pm$0.2}} &
  3.9$\pm$0.2 &
  { 74.1$\pm$0.3} \\
Chaudhry et al. \cite{chaudhry2019tiny} &
  15.7$\pm$0.3 &
  \multicolumn{1}{c|}{{ 66.1$\pm$1.3}} &
  21.3$\pm$0.5 &
  \multicolumn{1}{c|}{{ 59.3$\pm$0.9}} &
  28.8$\pm$0.8 &
  { 60.0$\pm$1.6} &
  4.7$\pm$0.5 &
  \multicolumn{1}{c|}{{ 68.2$\pm$2.8}} &
  10.1$\pm$0.7 &
  \multicolumn{1}{c|}{{ 66.2$\pm$0.8}} &
  11.7$\pm$0.2 &
  { 67.2$\pm$0.2} \\
Aljundi et al. \cite{aljundi2019online} &
  16.0$\pm$0.4 &
  \multicolumn{1}{c|}{{ 24.5$\pm$0.3}} &
  19.0$\pm$0.1 &
  \multicolumn{1}{c|}{{ 21.4$\pm$0.3}} &
  24.1$\pm$0.2 &
  { 21.0$\pm$0.1} &
  6.1$\pm$0.5 &
  \multicolumn{1}{c|}{{ 61.1$\pm$3.2}} &
  11.7$\pm$0.2 &
  \multicolumn{1}{c|}{{ 60.4$\pm$0.5}} &
  13.5$\pm$0.2 &
  { 59.5$\pm$0.3} \\
Aljundi et al. \cite{aljundi2019gradient} &
  11.1$\pm$0.2 &
  \multicolumn{1}{c|}{{ 73.4$\pm$4.2}} &
  13.3$\pm$0.5 &
  \multicolumn{1}{c|}{{ 69.3$\pm$3.1}} &
  17.4$\pm$0.1 &
  { 70.9$\pm$2.9} &
  3.3$\pm$0.5 &
  \multicolumn{1}{c|}{{ 72.8$\pm$1.2}} &
  10.0$\pm$0.2 &
  \multicolumn{1}{c|}{{ 72.6$\pm$0.4}} &
  10.5$\pm$0.2 &
  { 71.5$\pm$0.2} \\
Hou et al. \cite{hou2019learning} &
  8.6$\pm$1.3 &
  \multicolumn{1}{c|}{{ 60.0$\pm$0.1}} &
  19.5$\pm$0.7 &
  \multicolumn{1}{c|}{{ 47.5$\pm$0.9}} &
  16.9$\pm$0.5 &
  { 44.3$\pm$0.7} &
  7.6$\pm$0.5 &
  \multicolumn{1}{c|}{{ 46.4$\pm$0.7}} &
  9.6$\pm$0.7 &
  \multicolumn{1}{c|}{{ 42.2$\pm$0.9}} &
  12.5$\pm$0.7 &
  { 37.6$\pm$0.7} \\
Wu et al. \cite{wu2019large} &
  21.2$\pm$0.3 &
  \multicolumn{1}{c|}{{ 40.2$\pm$0.4}} &
  36.1$\pm$1.3 &
  \multicolumn{1}{c|}{{ 30.9$\pm$0.7}} &
  42.5$\pm$1.2 &
  { 18.7$\pm$0.5} &
  10.2$\pm$0.9 &
  \multicolumn{1}{c|}{{ 43.5$\pm$0.5}} &
  18.9$\pm$0.3 &
  \multicolumn{1}{c|}{{ 32.9$\pm$0.5}} &
  25.2$\pm$0.6 &
  { 24.9$\pm$0.4} \\
Prabhu et al. \cite{prabhu2020gdumb} &
  17.1$\pm$0.4 &
  \multicolumn{1}{c|}{{ 16.7$\pm$0.5}} &
  25.1$\pm$0.2 &
  \multicolumn{1}{c|}{{ 17.6$\pm$0.2}} &
  38.6$\pm$0.5 &
  { 16.8$\pm$0.4} &
  12.6$\pm$0.1 &
  \multicolumn{1}{c|}{{ \textbf{15.9$\pm$0.5}}} &
  12.7$\pm$0.3 &
  \multicolumn{1}{c|}{{ \textbf{14.6$\pm$0.3}}} &
  15.7$\pm$0.2 &
  { \textbf{11.7$\pm$0.2}} \\
Buzzega et al. \cite{buzzega2020dark} &
  15.3$\pm$0.2 &
  \multicolumn{1}{c|}{{ 43.4$\pm$0.2}} &
  19.7$\pm$1.5 &
  \multicolumn{1}{c|}{{ 44.0$\pm$1.9}} &
  27.0$\pm$0.7 &
  { 25.8$\pm$3.5} &
  4.5$\pm$0.3 &
  \multicolumn{1}{c|}{{ 67.2$\pm$1.7}} &
  10.1$\pm$0.3 &
  \multicolumn{1}{c|}{{ 63.6$\pm$0.3}} &
  17.6$\pm$0.5 &
  { 55.2$\pm$0.7} \\
Zhu et al. \cite{zhu2021class} &
  18.2$\pm$1.2 &
  \multicolumn{1}{c|}{{ 24.6$\pm$0.6}} &
  19.7$\pm$0.5 &
  \multicolumn{1}{c|}{{ 12.5$\pm$0.7}} &
  22.4$\pm$0.2 &
  { 20.0$\pm$0.5} &
  5.5$\pm$0.7 &
  \multicolumn{1}{c|}{{ 65.5$\pm$0.7}} &
  8.1$\pm$1.2 &
  \multicolumn{1}{c|}{{ 60.1$\pm$0.5}} &
  11.6$\pm$0.4 &
  { 57.6$\pm$1.1} \\
Cha et al. \cite{cha2021co2l} &
  17.1$\pm$0.4 &
  \multicolumn{1}{c|}{{ 16.9$\pm$0.4}} &
  24.2$\pm$0.2 &
  \multicolumn{1}{c|}{{ 16.6$\pm$0.6}} &
  32.2$\pm$0.5 &
  { 9.9$\pm$0.7} &
  10.1$\pm$0.2 &
  \multicolumn{1}{c|}{{ 60.5$\pm$0.5}} &
  15.8$\pm$0.4 &
  \multicolumn{1}{c|}{{ 52.5$\pm$0.9}} &
  22.5$\pm$1.2 &
  { 42.5$\pm$0.8} \\
Mai et al. \cite{mai2021supervised} &
  27.3$\pm$0.4 &
  \multicolumn{1}{c|}{{ 17.5$\pm$0.2}} &
  30.8$\pm$0.5 &
  \multicolumn{1}{c|}{{ 11.6$\pm$0.5}} &
  36.5$\pm$0.3 &
  { 5.6$\pm$0.4} &
  12.6$\pm$1.1 &
  \multicolumn{1}{c|}{{ 19.4$\pm$0.3}} &
  18.2$\pm$0.1 &
  \multicolumn{1}{c|}{{ {15.4$\pm$0.3}}} &
  21.1$\pm$1.1 &
  { 14.9$\pm$0.7} \\
Mittal et al. \cite{mittal2021essentials} &
  18.5$\pm$0.3 &
  \multicolumn{1}{c|}{{ 16.7$\pm$0.5}} &
  19.1$\pm$0.4 &
  \multicolumn{1}{c|}{{ 16.1$\pm$0.3}} &
  20.5$\pm$0.3 &
  { 17.5$\pm$0.2} &
  5.6$\pm$0.9 &
  \multicolumn{1}{c|}{{ 59.4$\pm$0.3}} &
  7.0$\pm$0.5 &
  \multicolumn{1}{c|}{{ 56.2$\pm$1.3}} &
  15.2$\pm$0.5 &
  { 48.9$\pm$0.6} \\

Ahn et al. \cite{ahn2021ss} &
  26.0$\pm$0.1 &
  \multicolumn{1}{c|}{{ 40.1$\pm$0.5}} &
  33.1$\pm$0.5 &
  \multicolumn{1}{c|}{{ 33.9$\pm$1.2}} &
  39.5$\pm$0.4 &
  { 21.7$\pm$0.8} &
  9.6$\pm$0.7 &
  \multicolumn{1}{c|}{{ 44.4$\pm$0.7}} &
  15.2$\pm$1.5 &
  \multicolumn{1}{c|}{{ 36.6$\pm$0.7}} &
  21.1$\pm$0.1 &
  { 29.0$\pm$0.7} \\
Shim et al. \cite{shim2021online} &
  16.4$\pm$0.3 &
  \multicolumn{1}{c|}{{ 25.0$\pm$0.2}} &
  12.2$\pm$1.9 &
  \multicolumn{1}{c|}{{ 12.2$\pm$1.9}} &
  27.1$\pm$0.3 &
  { 13.2$\pm$0.1} &
  5.3$\pm$0.3 &
  \multicolumn{1}{c|}{{ 65.7$\pm$0.7}} &
  8.2$\pm$0.2 &
  \multicolumn{1}{c|}{{ 64.2$\pm$0.2}} &
  10.3$\pm$0.4 &
  { 62.2$\pm$0.1} \\
Caccia et al. \cite{caccia2022new} &
  16.1$\pm$0.4 &
  \multicolumn{1}{c|}{{ 51.5$\pm$0.8}} &
  17.6$\pm$0.5 &
  \multicolumn{1}{c|}{{ 49.2$\pm$0.5}} &
  22.6$\pm$0.1 &
  { 38.7$\pm$0.6} &
  5.4$\pm$0.2 &
  \multicolumn{1}{c|}{{ 47.4$\pm$0.5}} &
  7.1$\pm$0.5 &
  \multicolumn{1}{c|}{{ 43.2$\pm$0.3}} &
  10.1$\pm$0.4 &
  { 41.0$\pm$0.5} \\
Guo et al. \cite{guo2022online} &
  28.1$\pm$0.3 &
  \multicolumn{1}{c|}{{ {12.2$\pm$0.3}}} &
  35.0$\pm$0.4 &
  \multicolumn{1}{c|}{{ {8.5$\pm$0.3}}} &
  42.4$\pm$0.5 &
  { \textbf{4.5$\pm$0.3}} &
  15.7$\pm$0.2 &
  \multicolumn{1}{c|}{{ 23.5$\pm$1.9}} &
  21.2$\pm$0.4 &
  \multicolumn{1}{c|}{{ 21.0$\pm$0.3}} &
  27.0$\pm$0.3 &
  { 18.6$\pm$0.5} \\
Wei et al. \cite{wei2023online} &
  30.0$\pm$0.4 &
  \multicolumn{1}{c|}{{ \textbf{10.4$\pm$0.5}}} &
  35.9$\pm$0.6 &
  \multicolumn{1}{c|}{{ \textbf{6.1$\pm$0.6}}} &
  41.3$\pm$0.5  &
  { {5.3$\pm$0.6}} &
  16.9$\pm$0.4 &
  \multicolumn{1}{c|}{{ {17.4$\pm$0.4}}} &
  22.1$\pm$0.4 &
  \multicolumn{1}{c|}{{ 16.8$\pm$0.4}} &
 29.8$\pm$0.5 &
  { 14.6$\pm$0.3} \\
Guo et al. \cite{guo2023dealing} &
  31.4$\pm$0.2 &
  \multicolumn{1}{c|}{{ 33.2$\pm$0.6}} &
  39.7$\pm$0.6 &
  \multicolumn{1}{c|}{{ 22.8$\pm$0.4}} &
  49.7$\pm$0.2 &
  { 8.7$\pm$0.3} &
  18.4$\pm$0.4 &
  \multicolumn{1}{c|}{{ 35.5$\pm$0.3}} &
  26.0$\pm$0.2 &
  \multicolumn{1}{c|}{{ 25.8$\pm$0.4}} &
  33.2$\pm$0.4 &
  { 16.9$\pm$0.6} \\
Yan et al. \cite{yan2024orchestrate} &
  \textbf{37.4$\pm$0.3} &
  \multicolumn{1}{c|}{{ 34.7$\pm$0.3}} &
  {\textbf{47.0$\pm$0.4}} &
  \multicolumn{1}{c|}{{ 23.6$\pm$0.4}} &
  {\textbf{55.6$\pm$0.4}} &
  { 12.7$\pm$0.4} &
  {\textbf{21.4$\pm$0.4}} &
  \multicolumn{1}{c|}{{ 40.6$\pm$0.6}} &
  {\textbf{29.8$\pm$0.5}} &
  \multicolumn{1}{c|}{{ 26.3$\pm$0.5}} &
  {\textbf{39.3$\pm$0.8}} &
  { 13.9$\pm$0.6} \\ \bottomrule
\end{tabular}%
}
\label{tab:OCL}
\end{table*}

\subsection{Image Classification (Online Continual Learning)}

Online continual learning refers to a challenging sub-setting of continual learning that allows each data instance to be passed to the model only once. Over this setting, following \cite{yan2024orchestrate}, to make a more thorough comparison over numerous different continual learning methods, we take both average accuracy ($A_q$) and average forgetting ($F_q$) as the evaluation metrics, and we perform comparison over two datasets including CIFAR-100 \cite{krizhevsky2009learning} and Tiny-ImageNet \cite{le2015tiny}.

\noindent\textbf{CIFAR-100} is a 100-classes image classification dataset. Following \cite{guo2022online,guo2023dealing,wei2023online,yan2024orchestrate}, over the evaluation of online continual learning methods, this dataset is split into 10 tasks. Besides, following \cite{guo2022online,guo2023dealing,yan2024orchestrate}, the allowed memory size is set to 1,000, 2,000, and 5,000. 

\noindent\textbf{Tiny-ImageNet} is a 200-classes image classification dataset. This dataset is split into 100 tasks during evaluation following \cite{guo2022online,guo2023dealing,wei2023online,yan2024orchestrate}. 
Moreover, over this dataset, the allowed memory size is set to 2,000, 4,000, and 10,000 following \cite{guo2022online,guo2023dealing,yan2024orchestrate}. 

\noindent\textbf{Experimental results.} We report results under the online continual learning setting in Tab.~\ref{tab:OCL}. As shown, w.r.t. the average accuracy metric, the recent method proposed by Yan et al. \cite{yan2024orchestrate} outperforms other continual learning methods consistently. Yet, as for the average forgetting metric, across different settings and datasets, the method that achieves state-of-the-art performance varies.

\begin{table*}
\centering
\caption{Performance comparison among different methods under the few-shot class incremental learning setting.}
\resizebox{\textwidth}{!}{%
\begin{tabular}{r|cccccccccccc}
\toprule
Dataset &
  \multicolumn{12}{c}{CUB200} \\ \midrule
\multirow{2.5}{*}{Evaluation Metric} & \multicolumn{11}{c|}{Accuracy (\%) $\uparrow$} & \multirow{2.5}{*}{PD (\%) $\downarrow$} \\ \cmidrule{2-12}
& Task 0 & Task 1 & Task 2 & Task 3 & Task 4 & Task 5 & Task 6 & Task 7 & Task 8 & Task 9 & \multicolumn{1}{c|}{{Task 10}} & \\
\midrule \midrule
Zhang et al. \cite{zhang2020deepemd} & 75.35 & 70.69 & 66.68 & 62.34 & 59.76 & 56.54 & 54.61 & 52.52 & 50.73 & 49.20 & \multicolumn{1}{c|}{{47.60}} & 27.75 \\
Chi et al. \cite{chi2022metafscil} & 75.90 & 72.41 & 68.78 & 64.78 & 62.96 & 59.99 & 58.30 & 56.85 & 54.78 & 53.82 & \multicolumn{1}{c|}{{52.64}} & 23.26 \\
Zou et al. \cite{zou2022margin} & 79.57 & 76.07 & 72.94 & 69.82 & 67.80 & 65.56 & 63.94 & 62.59 & 60.62 & 60.34 & \multicolumn{1}{c|}{{59.58}} & 19.99 \\
Kang et al. \cite{kang2023on} & 78.07 & 74.58 & 71.37 & 67.54 & 65.37 & 62.60 & 61.07 & 59.37 & 57.53 & 57.21 & \multicolumn{1}{c|}{{56.75}} & 21.32 \\
Kim et al. \cite{kim2023warping} & 77.74 & 74.15 & 70.82 & 66.90 & 65.01 & 62.64 & 61.40 & 59.86 & 57.95 & 57.77 & \multicolumn{1}{c|}{{57.01}} & 20.73 \\
Zhuang et al. \cite{zhuang2023gkeal} & 78.88 & 75.62 & 72.32 & 68.62 & 67.23 & 64.26 & 62.98 & 61.89 & 60.20 & 59.21 & \multicolumn{1}{c|}{{58.67}} & 20.21 \\
Yang et al. \cite{yang2023neural} & 80.45 & 75.98 & 72.30 & 70.28 & 68.17 & 65.16 & 64.43 & 63.25 & 60.66 & 60.01 & \multicolumn{1}{c|}{{59.44}} & 21.01 \\
Zou et al. \cite{zou2024compositional} & \textbf{80.94} & \textbf{77.51} & \textbf{74.34} & \textbf{71.00} & \textbf{68.77} & \textbf{66.41} & \textbf{64.85} & \textbf{63.92} & \textbf{62.12} & \textbf{62.10} & \multicolumn{1}{c|}{{\textbf{61.17}}} & \textbf{19.77} \\
\bottomrule
\end{tabular}%
}
\label{tab:FSCIL}
\end{table*}

\subsection{Image Classification (Few-Shot Class Incremental Learning)}

Few-shot class incremental learning refers to another sub-setting of continual learning that can be particularly useful in practical applications. In this setting, for each new task, only a few shots of data instances would be given for each new class. Over this setting, we follow \cite{zou2024compositional} in its way of comparing recent methods. Specifically, we measure the accuracy after completing the learning process of each new task. Besides, we also use another metric named performance dropping rate (PD) \cite{zhang2021few}. We here perform a comparison over the CUB200 dataset \cite{wah2011caltech}.

\noindent\textbf{CUB200} is a 200-classes fine-grained image classification dataset. Following \cite{tao2020few,zhang2021few}, this dataset is first split into 100 base training classes and 100 new classes. After that, the 100 new classes are further utilized to define 10 continual learning tasks, where each task is 10-way 5-shot. For fair comparison among different methods on this dataset, following \cite{zou2024compositional}, ResNet18 is consistently used as the backbone for all methods.

\noindent\textbf{Experimental results.} We report results under the few-shot class incremental learning setting in Tab.~\ref{tab:FSCIL}. As shown, under this setting and on the CUB200 dataset, Zou et al. \cite{zou2024compositional} consistently outperforms all other methods on both the accuracy and performance dropping rate metrics.

\begin{table*}
\centering
\caption{Performance comparison among different methods under the class-incremental semantic segmentation setting in mIoU (\%).}
\resizebox{\textwidth}{!}{%
\begin{tabular}{r|cccccccccccc}
\toprule
Dataset &
  \multicolumn{12}{c}{ADE20K} \\ \midrule
\multirow{2.5}{*}{Setting} 
& \multicolumn{4}{c|}{100-50} 
& \multicolumn{4}{c|}{100-10} 
& \multicolumn{4}{c}{100-5} \\ \cmidrule{2-13}
& Base & New & All & \multicolumn{1}{c|}{{Average}}
& Base & New & All & \multicolumn{1}{c|}{{Average}} 
& Base & New & All & Average \\
\midrule \midrule
Cermelli et al. \cite{cermelli2020modeling}     
& 40.50                                         & 17.20                                         & 32.80                                         & \multicolumn{1}{c|}{37.30}             
& 38.30             & 11.30             & 29.20             & \multicolumn{1}{c|}{35.10}             
& 36.00             & 5.70              & 26.00             & 32.70             \\
Michieli et al. \cite{michieli2021continual}     
& 40.52                                         & 17.17                                         & 32.79                                         & \multicolumn{1}{c|}{37.31}            
& 37.26             & 12.13             & 28.94             & \multicolumn{1}{c|}{34.48}             
& 33.02             & 10.63             & 25.61             & 33.07             \\
Douillard et al. \cite{douillard2021plop}       
& 41.87                                         & 14.89                                         & 32.94                                         & \multicolumn{1}{c|}{{37.39}}             
& 40.48             & 13.61             & 31.59             & \multicolumn{1}{c|}{{36.64}}             
& 35.72             & 12.18             & 27.93             & 35.10             \\
Phan et al. \cite{phan2022class}          & 41.55                                         & 19.16                                         & 34.14                                         & \multicolumn{1}{c|}{{38.43}}             
& 38.96             & 21.28    & 33.11 & \multicolumn{1}{c|}{{37.47}} 
& 36.06             & 16.38    & 29.54             & 36.49             \\
Zhang et al. \cite{zhang2022representation}  & 42.30                                         & 18.80                                         & 34.50                                         & \multicolumn{1}{c|}{{---}}               
& 39.30             & 17.60             & 32.10             &\multicolumn{1}{c|}{{---}} 
& 38.50             & 11.50             & 29.60             & ---               \\
Cermelli et al. \cite{cermelli2023comformer}  & 44.70                             & \textbf{26.20}                                & 38.40                             & \multicolumn{1}{c|}{{41.20}} 
& 40.60 & 15.60             & 32.30             & \multicolumn{1}{c|}{{37.40}}             
& 39.50 & 13.60             & 30.90 & 36.50 \\
Zhu et al. \cite{zhu2023continual} 
& 44.06 & 24.96 & 37.74 & \multicolumn{1}{c|}{{---}} 
& \textbf{43.88} & \textbf{25.14} & \textbf{37.67} & \multicolumn{1}{c|}{{---}} 
& \textbf{43.35} & \textbf{18.53} & \textbf{35.13} & --- \\
Chen et al. \cite{chen2024saving}
& 42.40 & 24.20 & 36.40 & \multicolumn{1}{c|}{{---}} 
& 42.00 & 20.60 & 34.90 & \multicolumn{1}{c|}{{---}} 
& --- & --- & --- & --- \\
Kim et al. \cite{kim2024eclipse}
& 45.00 & 21.70 & 37.10 & \multicolumn{1}{c|}{{---}} 
& 43.40 & 17.40 & 34.60 & \multicolumn{1}{c|}{{---}} 
& 43.30 & 16.30 & 34.20 & --- \\
Gong et al. \cite{gong2024continual}                                
& \textbf{45.73}                                & 26.02                             & \textbf{39.20}                                & \multicolumn{1}{c|}{{\textbf{41.62} }}    
& 42.32    & 18.42 & 34.41    & \multicolumn{1}{c|}{{\textbf{38.41} }}    
& 40.82    & 15.83 & 32.55    & \textbf{38.64}    \\

\bottomrule
\end{tabular}%
}
\label{tab:CSS}
\end{table*}

\subsection{Semantic Segmentation}

Besides image classification, recently, numerous different methods have also focused on the class-incremental setting over the semantic segmentation task. Over this class-incremental semantic segmentation setting, we compare recent methods in the same way as the recent method \cite{gong2024continual}. Specifically, we measure the mean interaction over union (mIoU) metric. Here, we perform a comparison over the ADE20K dataset \cite{zhou2017scene}.

\noindent\textbf{ADE20K} is a large-scale semantic segmentation dataset. It contains 150 classes. On this dataset, following  \cite{gong2024continual}, we perform comparison over different methods under three different settings. Specifically, all these different settings use 100 out of 150 classes in their base training. Then for the remaining 50 classes, the \textit{100-50} forms them into just 1 continual learning task, the \textit{100-10} setting uses them to form 5 tasks, and the \text{100-5} setting forms 10 tasks. Then for each of the three settings, following \cite{gong2024continual}, we report the mIoU calculated over the 100 base training classes (\textit{base}), the mIoU calculated over the 50 new classes (\textit{new}), the mIoU calculated over all classes (\textit{all}), and the average mIoU of all learning steps (\textit{average}).

\noindent\textbf{Experimental results.} In Tab.~\ref{tab:CSS}, we report the results of different class-incremental semantic segmentation methods. As shown, across different settings, the method that achieves state-of-the-art performance varies.

\begin{table*}
\centering
\caption{Performance comparison among different methods under the class-incremental object detection setting in AP (\%).}
\resizebox{\textwidth}{!}{%
\begin{tabular}{r|cccccccccccc}
\toprule
Dataset &
  \multicolumn{12}{c}{MS COCO 2017} \\ \midrule
\multirow{2.5}{*}{Setting} 
& \multicolumn{6}{c|}{40 + 40} 
& \multicolumn{6}{c}{70 + 10} \\ \cmidrule{2-13}
& $AP_{[0.5-0.95]}$ & $AP_{0.5}$ & $AP_{0.75}$ & $AP_S$ & $AP_M$ & \multicolumn{1}{c|}{$AP_L$}
& $AP_{[0.5-0.95]}$ & $AP_{0.5}$ & $AP_{0.75}$ & $AP_S$ & $AP_M$ & $AP_L$  \\
\midrule \midrule
Li and Hoiem \cite{li2017learning} 
& 17.2 & 25.4 & 18.6 & 7.9 & 18.4 & \multicolumn{1}{c|}{24.3} 
& 7.1 & 12.4 & 7.0 & 4.8 & 9.5 & 10.0 \\ 
Li et al. \cite{li2019rilod} 
& 29.9 & 45.0 & 32.0 & 15.8 & 33.0 & \multicolumn{1}{c|}{40.5} 
& 24.5 & 37.9 & 25.7 & 14.2 & 27.4 & 33.5 \\ 
Peng et al. \cite{peng2021sid} 
& 34.0 & 51.4 & 36.3 & 18.4 & 38.4 & \multicolumn{1}{c|}{44.9} 
& 32.8 & 49.0 & 35.0 & 17.1 & 36.9 & 44.5 \\ 
Feng et al. \cite{feng2022overcoming}  
& 36.9 & 54.5 & 39.6 & 21.3 & 40.4 & \multicolumn{1}{c|}{47.5} 
& 34.9 & 51.9 & 37.4 & 18.7 & 38.8 & 45.5 \\ 
Liu et al. \cite{liu2023continual} 
& 42.0 & 60.1 & 45.9 & 24.0 & 45.3 & \multicolumn{1}{c|}{55.6} 
& 40.4 & 58.0 & 43.9 & 23.8 & 43.6 & 53.5 \\                                    
Kim et al. \cite{kim2024sddgr} 
& \textbf{43.0} & \textbf{62.1} & \textbf{47.1} & \textbf{24.9} & \textbf{46.9} & \multicolumn{1}{c|}{\textbf{57.0}}
& \textbf{40.9} & \textbf{59.5} & \textbf{44.8} & \textbf{23.9} & \textbf{44.7} & \textbf{54.0} \\ 
\bottomrule
\end{tabular}%
}
\label{tab:CSOD}
\end{table*}

\subsection{Object Detection}

In addition, as time passed by, continual (class-incremental) object detection has also received increasing research attention. Over this task, we conduct performance comparison in the same way as the recent object detection work \cite{kim2024sddgr}. In specific, the comparison is conducted on the MS COCO 2017 dataset \cite{lin2014microsoft}, and the mean average precision (AP) at different interactions over union thresholds and object sizes (i.e., $AP_{[0.5-0.95]}$, $AP_{0.5}$, $AP_{0.75}$, $AP_S$, $AP_M$, $AP_L$) is utilized as the evaluation metric.

\noindent\textbf{MS COCO 2017} is an 80-classes object detection dataset. On this set, following \cite{kim2024sddgr}, we conduct performance comparison under two different settings, in which the first setting has 40 base training classes and 40 new classes (40 + 40), whereas the setting has 70 base training classes and 10 new classes (70 + 10).

\noindent\textbf{Experimental results.} We report the results of numerous class-incremental object detection methods in Tab.~\ref{tab:CSOD}. As shown, over this task, across different report metrics, the most recent method \cite{kim2024sddgr} consistently outperforms all the other methods, showing its effectiveness.

\begin{table*}
\centering
\caption{Performance comparison among different methods under the continual (lifelong) person re-identification setting.}
\resizebox{\textwidth}{!}{%
\begin{tabular}{r|cccccccccccc|cccccccccccc}
\toprule
Training Order & \multicolumn{12}{c|}{Order 1} & \multicolumn{12}{c}{Order 2}\\ \midrule
Dataset 
& \multicolumn{2}{c}{Market-1501} 
& \multicolumn{2}{c}{CUHK-SYSU}
& \multicolumn{2}{c}{DukeMTMC-ReID}
& \multicolumn{2}{c}{MSMT17-V2}
& \multicolumn{2}{c|}{CUHK03}
& \multicolumn{2}{c|}{Average}
& \multicolumn{2}{c}{Market-1501} 
& \multicolumn{2}{c}{CUHK-SYSU}
& \multicolumn{2}{c}{DukeMTMC-ReID}
& \multicolumn{2}{c}{MSMT17-V2}
& \multicolumn{2}{c|}{CUHK03}
& \multicolumn{2}{c}{Average}
\\ \midrule
Evaluation Metric
& mAP & R@1
& mAP & R@1
& mAP & R@1
& mAP & R@1
& mAP & \multicolumn{1}{c|}{R@1}
& mAP & \multicolumn{1}{c|}{R@1}
& mAP & R@1
& mAP & R@1
& mAP & R@1
& mAP & R@1
& mAP & \multicolumn{1}{c|}{R@1}
& mAP & R@1
\\ \midrule \midrule
Li and Hoiem \cite{li2017learning}
&56.3&77.1&72.9&75.1&29.6&46.5&6.0&16.6&36.1&\multicolumn{1}{c|}{37.5}&40.2&50.6
&42.7&61.7&5.1&14.3&34.4&58.6&69.9&73.0&34.1&\multicolumn{1}{c|}{34.1}&37.2&48.4 \\
Tung and Mori \cite{tung2019similarity} 
&35.6&61.2&61.7&64.0&27.5&47.1&5.2&15.5&\textbf{42.2}&\multicolumn{1}{c|}{\textbf{44.3}}&34.4&46.4 
&28.5&48.5&3.7&11.5&32.3&57.4&62.1&65.0&\textbf{43.0}&\multicolumn{1}{c|}{\textbf{45.2}}&33.9&45.5\\
Zhao et al. \cite{zhao2021continual}
&58.0&78.2&72.5&75.1&28.3&45.2&6.0&15.8&37.4&\multicolumn{1}{c|}{39.8}&40.5&50.8
&43.5&63.1&4.8&13.7&35.0&59.8&70.0&72.8&34.5&\multicolumn{1}{c|}{36.8}&37.6&49.2 \\
Pu et al. \cite{pu2021lifelong}
&58.1&77.4&72.5&74.8&28.7&45.2&6.1&16.2&38.7&\multicolumn{1}{c|}{40.4}&40.8&50.8
&42.2&60.1&5.4&15.1&37.2&59.8&71.2&73.9&36.9&\multicolumn{1}{c|}{37.9}&38.6&49.4 \\
Sun and Mu \cite{sun2022patch}
&68.5&85.7&75.6&78.6&\textbf{33.8}&\textbf{50.4}&6.5&17.0&34.1&\multicolumn{1}{c|}{36.8}&43.7&\textbf{53.7}
&58.3&74.1&6.4&17.4&\textbf{43.2}&\textbf{67.4}&74.5&76.9&33.7&\multicolumn{1}{c|}{34.8}&43.2&54.1\\
Pu et al. \cite{pu2023memorizing}
&39.0&61.6&73.3&76.6&16.9&30.3&4.6&13.4&36.4&\multicolumn{1}{c|}{37.1}&34.0&43.8
&21.6&35.5&3.0&9.3&25.0&49.8&69.9&73.1&34.7&\multicolumn{1}{c|}{35.1}&30.8&40.6 \\
Cui et al. \cite{cui2024learning}
&\textbf{69.0}&\textbf{86.8}&\textbf{76.7}&\textbf{79.5}&33.2&48.6&\textbf{6.6}&\textbf{17.4}&35.6&\multicolumn{1}{c|}{36.2}&\textbf{44.2}&\textbf{53.7}
&\textbf{59.7}&\textbf{75.0}&\textbf{7.3}&\textbf{19.2}&42.4&66.5&\textbf{76.0}&\textbf{77.8}&37.8&\multicolumn{1}{c|}{39.3}&\textbf{44.7}&\textbf{55.6} \\
\bottomrule
\end{tabular}%
}
\label{tab:CSPR}
\end{table*}

\subsection{Person Re-identification}

Moreover, the exploration of continual learning in the person re-identification task recently has also gained research attention. On this continual (lifelong) person re-identification task, we follow \cite{pu2021lifelong,cui2024learning} in their way of comparing different methods. In specific, rather than performing evaluation inner a certain dataset, here in continual person re-identification, to evaluate a method, the method is continually trained in order over a set of different datasets, and two evaluation metrics including the mean average precision (mAP) and the Rank@1 accuracy (R@1) are reported over each dataset. 

Here, following \cite{pu2021lifelong,cui2024learning}, we use five datasets including Market-1501 \cite{zheng2015scalable}, CUHKSYSU \cite{xiao2016end}, DukeMTMC-ReID \cite{zheng2017unlabeled}, MSMT17-V2 \cite{wei2018person}
and CUHK03 \cite{li2014deepreid}. Two different training orders are then employed based on the above 5 datasets, including Market-1501 $\rightarrow$ CUHK-SYSU
$\rightarrow$ DukeMTMC-ReID $\rightarrow$ MSMT17-V2 $\rightarrow$ CUHK03 (Order 1), and DukeMTMC-ReID $\rightarrow$ MSMT17-V2 $\rightarrow$ Market-1501 $\rightarrow$ CUHK-SYSU $\rightarrow$ CUHK03 (Order 2).

\noindent\textbf{Experimental results.} In Tab.~\ref{tab:CSPR}, we report the results of different continual person re-identification methods. As shown, on average, the most recent method \cite{cui2024learning} consistently outperforms all other methods. Yet, over different datasets, the method that achieves state-of-the-art performance still varies.

\section{Discussion}

In this section, we briefly discuss some potential directions that could be further investigated. 

Firstly, while a group of methods has been proposed in several other computer vision tasks, currently, most of the existing continual learning methods still focus on only the image classification problem. 
Although continual learning in image classification is a valuable topic to be explored, successful applications of continual learning to other computer vision problems are valuable as well. As with different characteristics of different computer vision problems, a simple adaptation of methods proposed for image classification may not lead to satisfactory performance in other computer vision problems. 
For example, in video grounding, Jin et al. \cite{jin2020visually} pointed out that simple adaptation of ideas from image classification fails with this compositional phrases learning scenario of the language input. 
In VQA, Perez et al. \cite{perez2018film} pointed out that their model cannot preserve previously learned knowledge well after being trained continuously on objects with different colors. Greco et al. \cite{greco2019psycholinguistics} further pointed out that besides the color of the objects, question difficulty and order also affect the degree of catastrophic forgetting in the area of VQA, and directly adapting methods from image classification can only mitigate catastrophic forgetting to a small extent. 
Besides visual grounding and VQA, Parcalabescu et al. \cite{parcalabescu2020seeing} also showed the failure of pre-trained vision and language models in the continual learning scenario of image sentence alignment and counting and pointed out the potential catastrophic forgetting problem. 
Other than various visual and language tasks, in image semantic segmentation, as different classes can co-exist in the same image, the model outputs corresponding to different classes can affect each other. Hence, Cermelli et al. \cite{cermelli2020modeling} pointed out that simple adaptation of ideas from image classification which regards the previously learned tasks as background during the learning process of the new task can even encourage rather than mitigate catastrophic forgetting. Ideas that are not effective in image classification may become effective in other computer vision tasks. For example, in analogical reasoning, Hayes and Kanan \cite{hayes2021selective} showed that selective memory retrieval compared to random memory retrieval can effectively mitigate catastrophic forgetting. However, selective memory retrieval has been shown to be less effective in image classification \cite{chaudhry2018riemannian,hayes2020remind}. Hence, we believe that besides image classification, continual learning in other computer vision problems is also worthy to be further investigated.

Secondly, although a few works such as \cite{hung2019increasingly} have applied continual learning to multi-problem learning, almost all continual learning studies nowadays focus on developing problem-specific continual learning algorithms such as image classification or image semantic segmentation. As knowledge can be shared among different computer vision problems, e.g., face recognition, facial identification, and facial anti-spoofing, how to continuously learn such kind of shared information in a multi-problem learning setting is worth exploring.

Thirdly, a large proportion of the existing continual learning approaches focus on the fully supervised problem setup with no class overlap between different tasks. However, in real-world scenarios, it is more likely that data instances that come at different time steps are unlabelled and have common classes between each other. Hence, there still exists a gap between most existing continual learning problem setups and real-world scenarios. We believe that continual learning methods with other alternative problem setups including unsupervised learning or self-supervised learning also warrant further investigations.

Finally, many continual learning works focus on the evaluation of performance/accuracy. Although some of them force their methods to only use limited memory space, the computation complexity and resource assumption, which are important factors for practical applications, are often not evaluated in many continual learning works. Considering the application of continuous learning on lightweight devices, the exploration of lightweight continuous learning methods is also important.

\section{Conclusion}

Continual learning is an important complementary of typical batch learning in neural network training. Recently, it has attracted broad attention, especially in the computer vision area. In this paper, we have given a comprehensive overview of the recent advances of continual learning in computer vision with various techniques used and various subareas that the methods have been applied to. We also conduct performance comparisons over continual learning methods in different computer vision tasks and give a brief discussion on some potential future research directions.

%\backmatter
\section{Author contributions}

Haoxuan Qu: Conceptualization, Visualization, Writing - original draft. Hossein Rahmani: Supervision, Writing - review \& editing. Li Xu: Writing - review \& editing. Bryan Williams: Writing - review \& editing. Jun Liu: Conceptualization, Project administration, Supervision, Writing - review \& editing.

% \bmsection*{Acknowledgments}
% Not applicable.

\section{Acknowledgments and Financial disclosure}

This work is supported by AISG-100E-2020-065, the SUTD Project PIE-SGP-Al2020-02, and the TAILOR project funded by EU Horizon 2020 research and innovation programme under GA No 952215.

\section{Conflict of interest}

The authors declare no potential conflict of interest.

% acknowledgments part
% \begin{acknowledgements}
% Not applicable.  
% % Acknowledgments of people, grants, \etc. 
% % you can acknowledge any support given which is not covered by the author 
% % contribution or funding sections. 
% % This may include administrative or technical support.

% % Funding: Please add “No funding was received to assist with the preparation of this manuscript.” ; 
% % “No funding was received for conducting this study.” 
% % Or “This research was funded by [NAME OF FUNDER], grant number […]” ; 
% % This work was supported by […] (Grant numbers […] and […]). 
% % Check carefully that the details given are accurate and use the standard spelling 
% % of funding agency names at https://search.crossref.org/funding.
% \end{acknowledgements}

% BibTeX from reference.bib
%\bibliographystyle{sn-apacite}
\bibliographystyle{unsrt}
\bibliography{reference}

\end{document}